\documentclass[11pt, a4paper]{scrartcl}
\pdfoutput=1
\usepackage[dvipsnames]{xcolor}

\usepackage[top=2.cm, bottom=2.cm, left=2.cm, right=2.cm]{geometry}

\usepackage{times}

\usepackage{verbatim}

\usepackage{pifont}
    \newcommand{\cmark}{\ding{51}}%
    \newcommand{\xmark}{\ding{55}}%

\usepackage{url}

\usepackage{array}
    \newcolumntype{C}{>{$}c<{$}}
    \newcolumntype{L}{>{$}l<{$}}
    \newcolumntype{R}{>{$}r<{$}}
\usepackage{multirow}

\usepackage[
    natbib,
    backend=biber,
    style=authoryear-comp,
    backref=false,
    maxbibnames=99,
    giveninits=false,
    uniquelist=false,
    maxcitenames=2,
    doi=false,
    isbn=false,
    url=false,
    uniquename=false,
    sortcites,
]{biblatex}
\AtEveryBibitem{%
	\clearlist{address}
	\clearfield{pages}
	\clearfield{volume}
	\clearfield{date}
	\clearfield{eprint}
	\clearfield{isbn}
	\clearfield{issn}
	\clearlist{location}
	\clearfield{month}
	\clearfield{series}
	
	\ifentrytype{book}{}{%
		\clearlist{publisher}
		\clearname{editor}
	}
}
\usepackage[colorinlistoftodos,
           prependcaption,
           textsize=small,
           backgroundcolor=yellow,
           linecolor=lightgray,
           bordercolor=lightgray
           ]{todonotes}

\usepackage{glossaries-extra}

\usepackage{graphicx}
\usepackage{subfigure}
\usepackage{booktabs} %
\usepackage{amsmath, latexsym}
\usepackage{amsfonts}
\usepackage{amssymb}
\usepackage{mathtools}
\usepackage{amsthm}
\usepackage{bm}
\usepackage{mathtools}
\usepackage[normalem]{ulem}
\usepackage[most]{tcolorbox}
\usepackage{placeins}
\usepackage{enumitem}
\usepackage{bold-extra}
\usepackage{chngcntr}

\usepackage[T1]{fontenc}

\usepackage{listings}
\definecolor{backcolour}{rgb}{0.95,0.95,0.92}
\definecolor{codegreen}{rgb}{0,0.6,0}

\lstdefinestyle{myStyle}{
    backgroundcolor=\color{backcolour},   
    commentstyle=\color{codegreen},
    basicstyle=\ttfamily\footnotesize,
    breakatwhitespace=false,         
    breaklines=true,                 
    keepspaces=true,                 
    numbers=left,       
    numbersep=5pt,                  
    showspaces=false,                
    showstringspaces=false,
    showtabs=false,                  
    tabsize=2,
    keywordstyle=\color{blue},       %
}

\usepackage{xspace}
\usepackage{makecell}

\usepackage{standalone}
\usepackage{tikz}
\usepackage{pgfplots}
\usepackage{pgf}
\usetikzlibrary{calc}
\usetikzlibrary{positioning}
\usetikzlibrary{angles,quotes}
\usetikzlibrary{backgrounds}
\usetikzlibrary{external}
\usetikzlibrary{fit}
\usetikzlibrary{arrows}
\usetikzlibrary{arrows.meta}
\usetikzlibrary{shapes.symbols}
\usetikzlibrary{shadings}
\usetikzlibrary{shapes}
\usetikzlibrary{fadings}
\usetikzlibrary{bayesnet}
\usetikzlibrary{matrix}
\usetikzlibrary{snakes}
\usetikzlibrary{decorations.pathmorphing, patterns}
\pgfplotsset{compat=1.15}
\usepgfplotslibrary{groupplots}

\definecolor{C1}{HTML}{1F77B4}
\definecolor{C2}{HTML}{FF7F0E}
\definecolor{C3}{HTML}{2CA02C}
\definecolor{C4}{HTML}{D62728}
\definecolor{C5}{HTML}{9467BD}
\colorlet{C1light}{C1!70!white}
\colorlet{C2light}{C2!70!white}
\colorlet{C3light}{C3!70!white}
\colorlet{C4light}{C4!70!white}
\colorlet{C5light}{C5!70!white}
\colorlet{C1vlight}{C1!20!white}
\colorlet{C2vlight}{C2!20!white}
\colorlet{C3vlight}{C3!20!white}
\colorlet{C4vlight}{C4!20!white}
\colorlet{C5vlight}{C5!20!white}
\colorlet{linkcolor}{violet}
\colorlet{citecolor}{RedOrange}  %
\colorlet{urlcolor}{Aquamarine}

\usepackage{caption}
\DeclareCaptionFont{textsc}{\bfseries\selectfont}
\usepackage{hyperref}
    \newcommand\myshade{85}
    \hypersetup{
      linkcolor  = linkcolor!\myshade!black,
      citecolor  = citecolor!\myshade!black,
      urlcolor   = urlcolor!\myshade!black,
      colorlinks = true,
    }

\usepackage[capitalise, nameinlink]{cleveref}
    \Crefname{table}{Tab.}{Tabs.}
    \Crefname{appendix}{App.}{Apps.}
    \Crefname{section}{Sec.}{Secs.}
    \Crefname{equation}{Eq.}{Eqs.}
    
\usepackage{defns}
\newabbreviation{da}{DA}{data augmentation}
\newabbreviation{tta}{TTA}{test-time augmentation}

\theoremstyle{plain}

\theoremstyle{definition}

\theoremstyle{remark}

\DeclareFieldFormat{citehyperref}{%
  \DeclareFieldAlias{bibhyperref}{noformat}%
  \bibhyperref{#1}}

\DeclareFieldFormat{textcitehyperref}{%
  \DeclareFieldAlias{bibhyperref}{noformat}%
  \bibhyperref{%
    #1%
    \ifbool{cbx:parens}
      {\bibcloseparen\global\boolfalse{cbx:parens}}
      {}}}

\savebibmacro{cite}
\savebibmacro{textcite}

\renewbibmacro*{cite}{%
  \printtext[citehyperref]{%
    \restorebibmacro{cite}%
    \usebibmacro{cite}}}

\renewbibmacro*{textcite}{%
  \ifboolexpr{
    ( not test {\iffieldundef{prenote}} and
      test {\ifnumequal{\value{citecount}}{1}} )
    or
    ( not test {\iffieldundef{postnote}} and
      test {\ifnumequal{\value{citecount}}{\value{citetotal}}} )
  }
    {\DeclareFieldAlias{textcitehyperref}{noformat}}
    {}%
  \printtext[textcitehyperref]{%
    \restorebibmacro{textcite}%
    \usebibmacro{textcite}}}

\makeatletter
\def\paragraph{\@startsection{paragraph}{4}{\z@}{1.5ex plus
  0.5ex minus .2ex}{-1em}{\normalsize\bfseries}}
\makeatother

\usepackage{parskip}
\usepackage{microtype}

\newcommand{\alex}[1]{\textcolor{magenta}{}}
\newcommand{\mbnote}[1]{\textcolor{C1}{}}
\newcommand{\soham}[1]{\textcolor{C3}{}}

\newcommand{\razvan}[1]{\textcolor{olive}{}}

\begin{document}
\title{Regularising for invariance to data augmentation improves supervised learning}
\author{Aleksander Botev$^\dagger$, Matthias Bauer$^\dagger$, Soham De$^\dagger$ \\ \texttt{\{botev|msbauer|sohamde\}@deepmind.com}\\[0.25em] DeepMind, London, UK \\[0.5em] \normalsize{$^\dagger$all authors contributed equally to this work}}

\maketitle

\begin{abstract}

Data augmentation is used in machine learning to make the classifier invariant to label-preserving transformations.
Usually this invariance is only encouraged implicitly by including a single augmented input during training. However, several works have recently shown that using multiple augmentations per input can improve generalisation or can be used to incorporate invariances more explicitly.
In this work, we first empirically compare these recently proposed objectives that differ in whether they rely on explicit or implicit regularisation and at what level of the predictor they encode the invariances.
We show that the predictions of the best performing method are also the most similar when compared on different augmentations of the same input. Inspired by this observation, we propose an explicit regulariser that encourages this invariance on the level of individual model predictions. Through extensive experiments on CIFAR-100 and ImageNet we show that this explicit regulariser (i) improves generalisation and (ii) equalises performance differences between all considered objectives.
Our results suggest that objectives that encourage invariance on the level of the neural network itself generalise better than those that achieve invariance by averaging predictions of non-invariant models.

\end{abstract}

\section{Introduction}
\label{sec:introduction}

In supervised learning problems, we often have prior knowledge that the labels should be invariant or insensitive to certain transformations of the inputs; though it is often difficult to %
make the classifier explicitly invariant to them in a tractable way \citep{Niyogi1998-data-augmentation}. 
\Gls{da} is a widely used technique to incorporate such an inductive bias into the learning problem implicitly: it enlarges the training set with randomly transformed copies of the original data to encourage the classifier to be correct on larger parts of the input space.

When using \gls{da}, practitioners typically sample a single augmentation per image and minibatch during training. However, \citet{Fort2021-yq} recently showed that this can introduce a detrimental variance that slows down training. 
Instead, using \emph{several} augmentations per image in the same minibatch can reduce this variance and improve the classifier's generalisation performance by leveraging the useful bias of \glspl{da} more effectively \citep{hoffer2019augment, hendrycks2019augmix, touvron2021going, Fort2021-yq}.
This modified objective simply takes the average of the individual losses over different augmentations of the same input, which encourages the model function to produce similar predictions for every augmentation. 

\begin{figure}[t]
    \centering
    \includestandalone[mode=image]{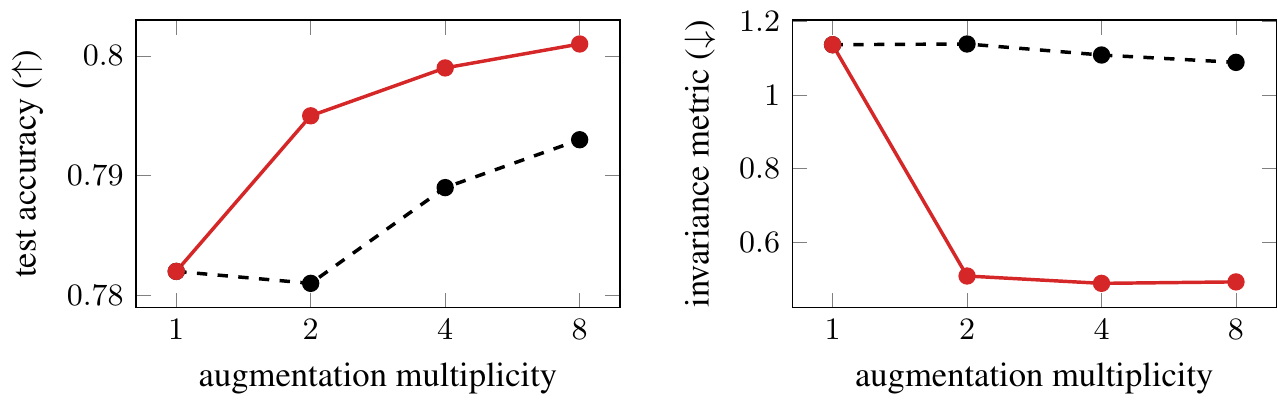}
    \vskip-0.5em
    \caption[]{
    Encouraging invariance to \glspl{da} through the proposed explicit regulariser
    (\protect\tikz[baseline=(current bounding box.base),inner sep=0pt]{\protect\draw [C4, thick] (0,3pt) -- +(0.5,0); \protect\node at (0.25, 3pt) {\textcolor{C4}{$\bullet$}};})
    consistently improves generalisation compared to a regular objective
    (\protect\tikz[baseline=(current bounding box.base),inner sep=0pt]{\protect\draw [black, thick, dashed] (0,3pt) -- +(0.5,0); \protect\node at (0.25, 3pt) {\textcolor{black}{$\bullet$}};})
    when sampling multiple augmentations per input during training. Values are with an NF-ResNet-101 on ImageNet.
    }
    \label{fig:introduction}
\end{figure}

The choice of averaging the individual losses
motivates the question of whether there are better ways of combining the model outputs when sampling multiple augmentations per input. In this direction,  \citet{Nabarro2021-bv} recently studied principled ways to incorporate data-augmentation in a Bayesian framework for neural networks by explicitly constraining the classifier outputs to be invariant to \gls{da} by averaging either (i) the post-softmax probabilities or (ii) the pre-softmax logits during training. 
This alternative of averaging the probabilities is often used to improve test performance by making ensemble predictions over different augmentations of the test input \citep{krizhevsky2012imagenet, simonyan2014very, szegedy2015going}. 
Interestingly, while \citet{Nabarro2021-bv} showed that both Bayesian-inspired approaches %
improved performance compared to sampling a single augmentation per input, they do not compare them to the baseline of averaging the losses, perhaps because this baseline does not correspond to a valid likelihood on the unaugmented dataset.

As a result, it remains an open question how and at what level practitioners should incorporate their prior knowledge of invariance in the model: (i) implicitly by averaging the losses over individual augmentations; or (ii) explicitly by constructing a model whose prediction is defined as the average over all augmentations, as is done when averaging the %
probabilities or the %
logits.
In this work we compare these two alternative perspectives. We show that averaging the losses during training leads to better generalisation performance, even when ensembling predictions for different augmentations at test time.
Additionally, we empirically show that the approach of averaging the losses makes the predictions of the neural network model for different augmentations significantly more similar than the other two methods.
Based on this observation, we conjecture that having more invariant predictions for individual data augmentations is the reason why averaging the losses generalises better.

To further investigate this hypothesis, we introduce a regulariser that explicitly encourages individual predictions for different \glspl{da} of the same input to be similar. Since we already use multiple \glspl{da} per input in our supervised setup, we can directly compare the predictive distributions of different \glspl{da} at no additional computational cost. Through extensive experiments, we show that this explicit regulariser consistently improves generalisation by making the model more invariant to \glspl{da} of the same input. As an example, in \cref{fig:introduction} we show the %
improvements when using the regulariser on an NF-ResNet-101 \citep{brock2021characterizing} trained on ImageNet when averaging the losses over multiple \glspl{da} (see \cref{sec:kl_loss} for details). Furthermore, we show that using this regulariser equalises the performance differences between averaging the losses vs.~averaging the logits or probabilities. 
These results corroborate our conjecture that encouraging invariance on the level of the network outputs is better than achieving invariance by averaging non-invariant models alone. 

\begin{figure*}[t]
    \centering
    \includestandalone[mode=image]{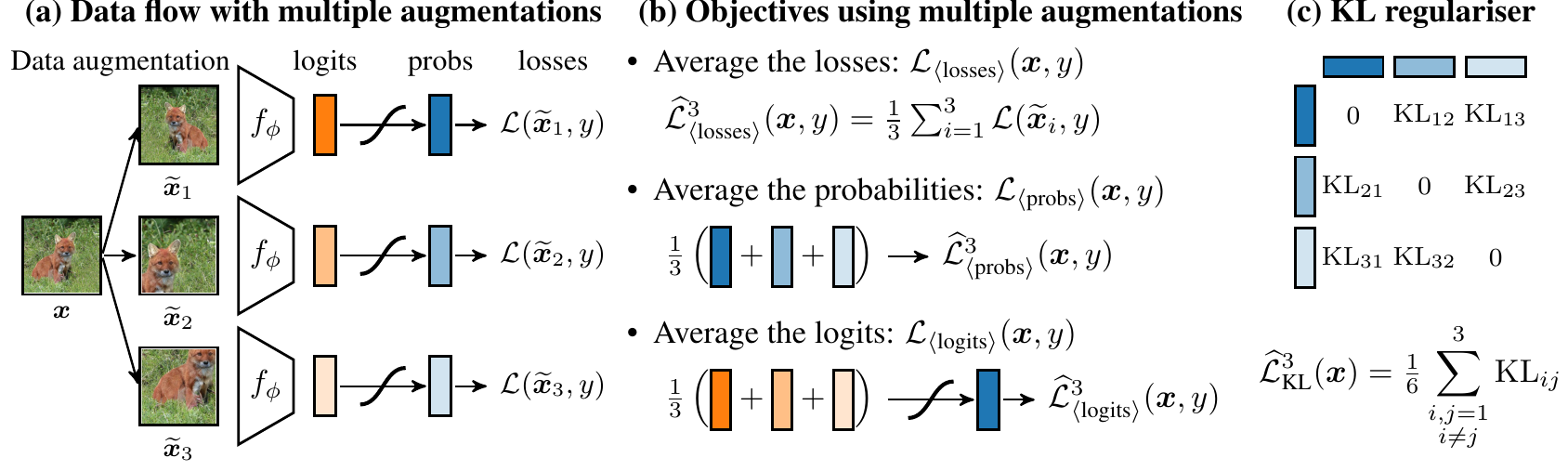}
    \vskip-0.25em
    \caption{
    We investigate three objectives that use multiple augmentations per input $\vx$: $\lossstd$ averages the losses of individual augmentations, while  $\lossprob$ and $\losslog$ average the probabilities and logits, respectively (\textit{left} and \textit{middle}). We further propose a regulariser that averages the KL-divergence between all pairs of predictive distributions
    over different augmentations of the same input
    (\textit{right}).
    }
    \label{fig:method}
\end{figure*}

\textbf{To summarise, our main contributions are:}
\begin{enumerate}[topsep=-2pt,itemsep=2pt,partopsep=0pt, parsep=0pt, leftmargin=*]
    \item We study the role of explicit and implicit invariance in supervised learning. 
    We find that a naive approach of averaging the losses outperforms Bayesian-inspired losses, with more invariant model predictions (\cref{sec:avg_methods}). 
    \item We propose a regulariser for supervised learning tasks that explicitly encourages predictions of different \glspl{da} for the same input to be similar to each other (\cref{sec:kl_loss}).
    \item We show that the regulariser improves generalisation for all base losses, equalising the performance differences between methods; this suggests that encouraging invariance on the level of the network outputs works better.
\end{enumerate}

\section{Background}
\label{sec:background}

In this work, we consider the supervised learning setting with input vectors $\vx\in\mathcal{X}\subseteq\mathbb{R}^d$ and corresponding labels $y\in\mathcal{Y}=\{1, \dots, C\}$. 
Let $\data=\left\{(\vx_n, y_n)\right\}_n^N$ denote the training set and $\fphi$ be a parametric (neural network) model with parameters $\vphi$.
The standard objective is to minimise the empirical negative log likelihood of the training data:
\begin{equation}
\label{eq:loss_orig}
\begin{aligned}
    \min_{\vphi} {} \quad & \expect{(\vx, y)\sim\data}{\mathcal{L}_{\vphi}(\vx, y)}, \\ \mathcal{L}_{\vphi}(\vx, y) &= -\log p(y \given g(\fphi(\vx))).
\end{aligned}
\end{equation}
Here, $g$ is an inverse link function (such as the logistic or softmax) that maps the output of the function $\fphi$ to a probability distribution over $\mathcal{Y}$, and $\mathcal{L}_{\vphi}(\vx, y)$ denotes the negative log likelihood objective for a single datapoint $(\vx, y)$.

\paragraph{Standard data augmentation (DA).} \gls{da} refers to the practice of enlarging the original dataset $\data$ by applying label-preserving transformations to its inputs $\vx$. The transformations are domain- and dataset-specific and hand-engineered to incorporate the right inductive biases such as invariances or symmetries \citep{Niyogi1998-data-augmentation}. For example, in image classification, we commonly use a combination of discrete transformations such as horizontal flips and continuous transformations such as shears or changes in brightness or contrast \citep{randAugment}.
Following \citet{Wilk2018-learning-invariances}, we call the distribution over all augmentations $\vxa$ for a given input $\vx$ its \emph{augmentation distribution} and denote it by $p(\vxa\given\vx)$. 
The per-input objective that corresponds to regular neural network training with \gls{da} is then given by:
\begin{equation}
    \label{eq:loss_std}
    \begin{aligned}
    \lossavglosses(\vx, y) & = \expect{\vxa \sim p(\vxa \given \vx)}{\varL_{\vphi}(\vxa, y)},%
    \end{aligned}
\end{equation}
and can be interpreted as averaging the losses for different augmentations of the same input.
The expectation in \cref{eq:loss_std} is commonly approximated by a single Monte Carlo sample, i.e., for each input in a minibatch we sample one augmentation. 
Several recent works have shown, however, that sampling a larger number of augmentations per input in a minibatch can be beneficial \citep{hoffer2019augment, hendrycks2019augmix, touvron2021going, Fort2021-yq}.
This phenomenon has been attributed to the observation that sampling multiple augmentations per input in a minibatch has the effect of reducing the variance arising from the \gls{da} procedure and empirically improves generalisation \citep{Fort2021-yq}.
This is in contrast to minibatch sampling where lowering the variance by increasing the batch size might reduce generalisation \citep{smith2020generalization}.

\paragraph{Bayesian-inspired \gls{da}.} 
The question of how best to incorporate \gls{da} in Bayesian deep learning has recently also received some attention, especially since the naive approach of enlarging the training set based on the number of augmentations results in overcounting the likelihood w.r.t.~the prior \citep{Nabarro2021-bv}. In this context, \citet{Wenzel2020-cold-posterior} noted that the data augmented objective in \cref{eq:loss_std} cannot be interpreted as a valid likelihood objective. 
\citet{Nabarro2021-bv} argued that \gls{da} should be viewed as nuisance variables that should be marginalised over, and proposed two Bayesian-inspired objectives that construct an invariant predictive distribution by integrating a non-invariant predictor over the augmentation distribution, where either the post-softmax probabilities or pre-softmax logits are averaged:
\begin{align}
    \lossavgprobs(\vx, y) &= -\log \expect{\vxa \sim p(\vxa \given \vx)}{p(y \given g(\fphi(\vxa)))}, \label{eq:loss_avg_probs}\\
    \lossavglogs(\vx, y) &= -\log p(y \given g(\expect{\vxa \sim p(\vxa \given \vx)}{\fphi(\vxa)})).
    \label{eq:loss_avg_logs}
\end{align}
This construction makes the predictive distribution \emph{explicitly} invariant to \glspl{da}, in contrast to the construction in \cref{eq:loss_std}.

\paragraph{Finite sample \gls{da} objectives.}
In this work, we first compare the above two alternate perspectives on \gls{da} to understand how best to incorporate invariance into the model.
Because exact marginalisation in \cref{eq:loss_std,eq:loss_avg_probs,eq:loss_avg_logs} is intractable, we approximate it with $K$ samples from the augmentation distribution giving rise to the following objectives:
\vspace{2mm}
\begin{tcolorbox}[enhanced,colback=white,%
    colframe=C1!75!black, attach boxed title to top right={yshift=-\tcboxedtitleheight/2, xshift=-.75cm}, title=Objectives using multiple augmentations, coltitle=C1!75!black, boxed title style={size=small,colback=white,opacityback=1, opacityframe=0}, size=title, enlarge top initially by=-\tcboxedtitleheight/2, left=-5pt, %
    ]
    \vspace{-0.75em}
\begin{align}
    \lossstdhat^K(\vx, y) &= \tfrac{1}{K}\textstyle\sum_{k=1}^K \varL\left(g(\fphi(\vxa_{k})), y\right) \label{eq:avg_losses_k}\\
    \lossprobhat^K(\vx, y) &= \varL\left(\tfrac{1}{K}\textstyle\sum_{k=1}^K g(\fphi(\vxa_{k})), y\right) \label{eq:avg_probs_k}\\
    \lossloghat^K(\vx, y) &= \varL\left(g\left(\tfrac{1}{K}\textstyle\sum_{k=1}^K\fphi(\vxa_{k})\right), y\right) \label{eq:avg_logs_k}
\end{align}
\end{tcolorbox}
Here the $\vxa_{k}, k\in\{1, \dots, K\}$, are the $K$ augmentations for an input $\vx$ independently sampled from $p(\vxa\given\vx)$. For $K=1$ the losses are equal.
In \cref{fig:method} (a) and (b) we illustrate these three objectives for $K=3$. 
See \cref{sec:loss_comparisons} for a brief analysis of how these three losses compare to each other, and how the finite sample versions of the objectives (\cref{eq:avg_losses_k,eq:avg_probs_k,eq:avg_logs_k}) relate to the original objectives (\cref{eq:loss_std,eq:loss_avg_probs,eq:loss_avg_logs}).

\paragraph{Test-time \gls{da} (TTA).} While so far we have discussed \gls{da} during \emph{training}, it is also common to employ it at test time by making ensemble predictions over multiple augmentations of the test input to boost performance \citep{krizhevsky2012imagenet, simonyan2014very, szegedy2015going}.
Note that for TTA, practitioners typically \emph{average the probabilities}  as in \cref{eq:avg_probs_k}, while for training practitioners typically \emph{average the losses} as in \cref{eq:avg_losses_k}. This inconsistency further motivates the question of whether it is also better to similarly ensemble the probabilities during training. 

\section[]{Experimentally comparing the two alternate perspectives on \protect\gls{da}}
\label{sec:avg_methods}
In \cref{sec:background}, we described two alternate perspectives on \gls{da} to incorporate invariance into the model: either implicitly by averaging the losses (\cref{eq:avg_losses_k}), or explicitly by averaging either the probabilities (\cref{eq:avg_probs_k}) or the logits (\cref{eq:avg_logs_k}). To better understand how these perspectives compare, in this section we perform a thorough experimental evaluation of the three objectives.
Specifically, we compare both generalisation performance and the amount of invariance induced by the three losses for varying number of augmentations per input per minibatch during training (which we call \emph{augmentation multiplicity} $\ktrain$). 
This evaluation extends previous limited results by \citet{Nabarro2021-bv}, who only compare $\lossprobhat^K$ and $\lossloghat^K$ for $\ktrain \in [1, 6]$ and do not investigate $\lossstdhat^K$ for $\ktrain$ larger than $1$, and \citet{Fort2021-yq}, who investigate $\lossstdhat^K$ but do not consider invariances. 

\paragraph{Experimental setup.}
Since the performance of models with batch normalisation depends strongly on the examples used to estimate the batch statistics \citep{hoffer2017train}, in the main paper we train on highly performant models that do not use batch normalisation, following \citet{Fort2021-yq}, to simplify our analysis (see \cref{app:resnet18} for more discussion).
We use the following networks: a WideResNet 16-4 \citep{wrn2016} with SkipInit initialisation \citep{de2020batch} for CIFAR-100 classification, and an NF-ResNet-101 \citep{brock2021characterizing} for ImageNet classification.
For both datasets we use standard random crops and random horizontal flips for \gls{da} following previous work \citep{wrn2016, brock2021characterizing}. See \cref{subsec:more_augs} for additional experiments with a wider range of data augmentations where we have similar findings.

We use augmentation multiplicities $\ktrain={1,2,4,8,16}$ for CIFAR-100 and $\ktrain={1,2,4,8}$ for ImageNet due to computational constraints. We also fix the total batch size, which implies that for $\ktrain > 1$ the number of unique images per single batch decreases proportionally and the total number of parameter updates for the same epoch budget increases \citep{Fort2021-yq}. Because the optimal epoch budget might change with the augmentation multiplicity \citep{Fort2021-yq}, for each experiment we run an extensive grid-search for the optimal learning rate and optimal epoch budget for every value of $\ktrain$ for the three objective functions. %
For evaluation, we compute the top-1 test accuracy both using standard central-crops as well as using test time augmentations (TTAs) with the number of augmentations set to $\ktest=16$ for CIFAR-100 and $\ktest=8$ for ImageNet. 
As a measure of the invariance of the predictions, we calculate the KL divergence between the predictive distributions of different augmentations of the same input (see \cref{sec:kl_loss} for a more detailed description of this measure).
We run each CIFAR-100 experiment $5$ times with different random seeds; for the ImageNet experiments we only use a single seed due to computational constraints.
For further details on the experimental setup, please refer to \cref{sec:app:experimental_details} . 

\begin{figure}[t]
    \centering
    \includestandalone[mode=image]{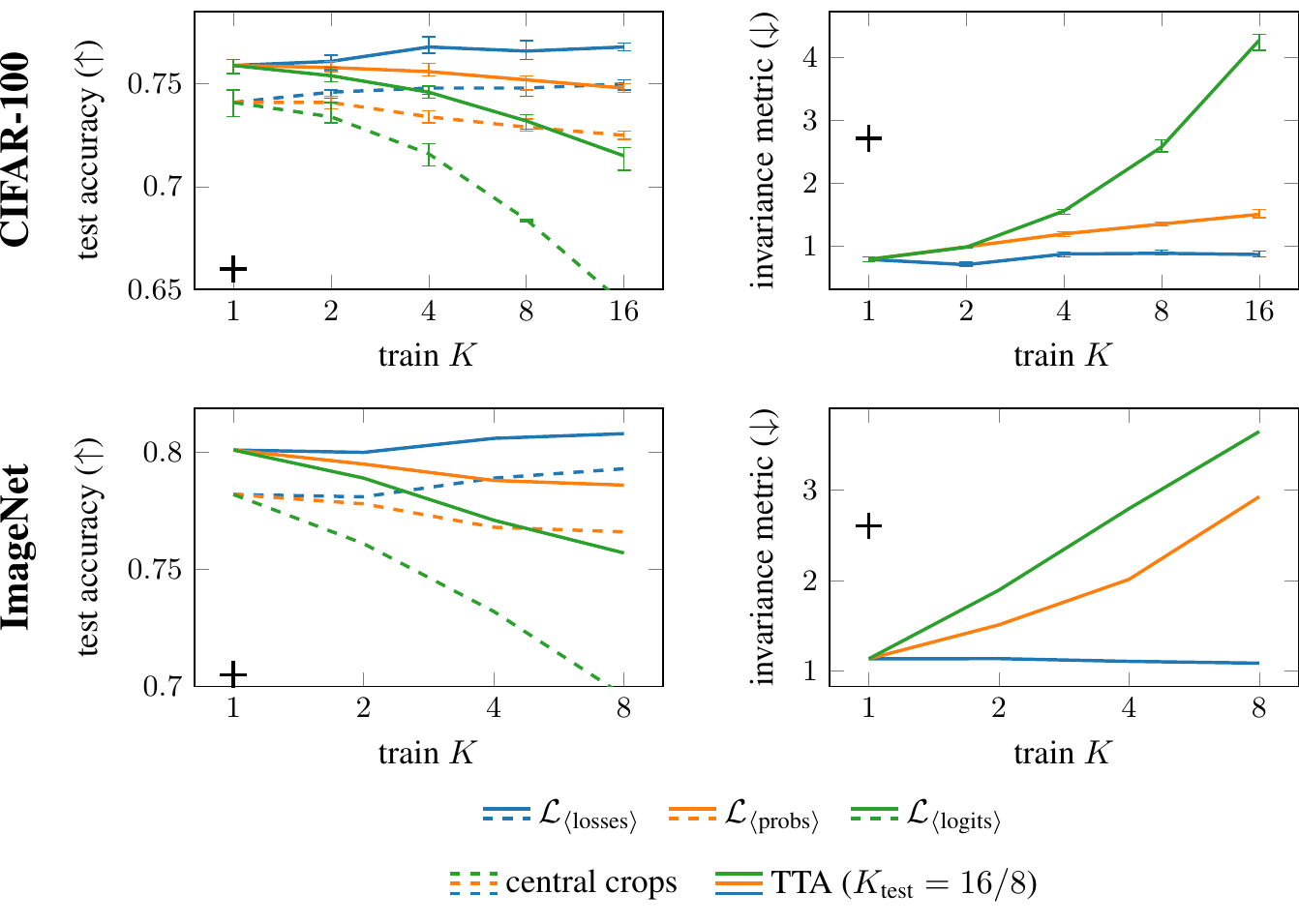}
    \vskip-0.5em
    \caption{
    The three losses differ in their generalisation performance \textit{(left)} as well as in how invariant the predictions are w.r.t.~data augmentations \textit{(right)}. %
    For reference, 
    \protect\tikz[baseline=(current bounding box.base),inner sep=0pt]{\protect\draw [black, thick] (0,0.075) -- +(0.25,0);\protect\draw [black, thick] (0.125,0.2) -- +(0,-0.25);}
    mark models trained without \protect\gls{da} (\protect\cref{eq:loss_orig}).
    }
    \label{fig:results_no_kl}
\end{figure}

\paragraph{Experimental results.}
Overall, the results are qualitatively the same on both datasets. 
For top-1 test accuracy we find that (see \cref{fig:results_no_kl} \textit{left} and \cref{tab:results}): (i) Using augmentations at test time is consistently at least $\sim2\%$ better than predictions on central crops; (ii) Averaging the losses ($\lossstd$) clearly performs better than averaging the probabilities ($\lossprob$) which performs better than averaging the logits ($\losslog$); (iii) While $\lossstd$ improves with larger augmentation multiplicity as reported by \citet{Fort2021-yq}, both Bayesian-inspired objectives, $\lossprob$ and $\losslog$, consistently \emph{degrade} in performance as $\ktrain$ increases.

When comparing how the invariance measure for different augmentations of the same input changes as we vary the augmentation multiplicity, we find that (see \cref{fig:results_no_kl} \textit{right}), while it stays relatively constant when averaging the losses, it markedly increases for both Bayesian-inspired losses as $\ktrain$ increases. We therefore conjecture that: the beneficial bias of \gls{da}, as discussed by \citet{Fort2021-yq}, is that it promotes invariance of the model to \gls{da}; the detrimental variance also introduced by \gls{da} can be mitigated by using larger augmentation multiplicities. This also explains why larger $\ktrain$ do not further reduce the invariance measure for $\lossstd$.
For completeness, we also show in \cref{fig:results_no_kl} that a network trained \emph{without} \gls{da} has a much higher invariance measure than networks trained with \gls{da} when $\ktrain = 1$.

\begin{figure*}[t]
    \centering
    \includestandalone[mode=image]{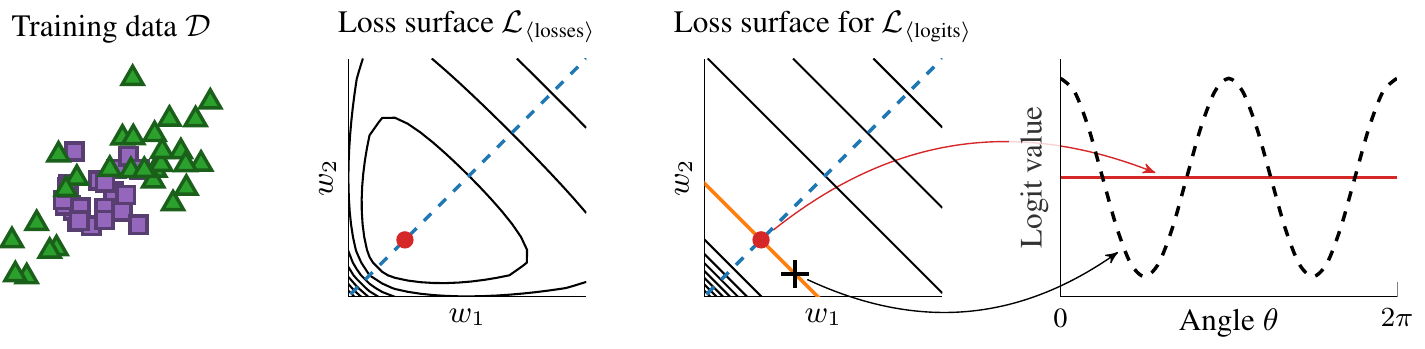}
    \vskip-0.5em
    \caption{
    Example why $\lossstd$ encourages invariance but $\losslog$ does not. We fit a $2$-parameter model to a binary classification problem with rotational symmetry \textit{(left)}. Setting $w_1=w_2$
    (\protect\tikz[baseline=(current bounding box.base),inner sep=0pt]{\protect\draw [C1, dashed, thick] (0,3pt) -- +(0.5,0);})
    corresponds to models with constant logit values regardless of the angle $\theta$. The loss surface for $\lossstd$ has a single optimum at an invariant model (\textcolor{C4}{$\bullet$}), whereas the loss surface for $\losslog$ is optimal anywhere along a line 
    (\protect\tikz[baseline=(current bounding box.base),inner sep=0pt]{\protect\draw [C2, thick] (0,0.25) -- +(0.3,-0.3);}), as it only constrains the \emph{average} logit value. 
    The model is invariant only for one point along the line
    (\textcolor{C4}{$\bullet$}),
    whereas for all other values 
    (e.g. \protect\tikz[baseline=(current bounding box.base),inner sep=0pt]{\protect\draw [black, thick] (0,0.075) -- +(0.25,0);\protect\draw [black, thick] (0.125,0.2) -- +(0,-0.25);})
    the logit value changes with the angle $\theta$. %
    }
    \label{fig:toy_example_no_kl}
\end{figure*}

\begin{table*}[ht]
	\centering
	\begin{tabular}{llc | CCCC}
		\toprule 
		\multirow{2}{*}{\textbf{Dataset}} & \multirow{2}{*}{\textbf{Network}} & \multirow{2}{*}{\textbf{KL regulariser}} & \multirow{2}{*}{$\ktrain=1$} & \multicolumn{3}{c}{$\ktrain=8 \text{ or }16$} \\
		&&&  &  \lossstd & \lossprob & \losslog \\\midrule
		CIFAR-100 &WRN 16-4 & \xmark & 0.741\pms{0.003}  & 0.750\pms{0.001} & 0.725\pms{0.001} & 0.643\pms{0.001}\\
		CIFAR-100 &WRN 16-4 & \cmark & 0.741\pms{0.003} & 0.759\pms{0.001} &  0.758\pms{0.002} &  0.759\pms{0.001}  \\\midrule
		ImageNet &NF-ResNet-101 & \xmark & 0.782 & 0.793 & 0.766 & 0.696  \\
		ImageNet &NF-ResNet-101 & \cmark & 0.782 & 0.801 & 0.804 & 0.804 \\ \midrule
        ImageNet & NFNet-F0 & \xmark & 0.803 & 0.813 & 0.788 & 0.717 \\
		ImageNet & NFNet-F0 & \cmark & 0.803 & 0.819 & 0.822 & 0.819\\
		\bottomrule
	\end{tabular}
	\caption[]{Top-1 test accuracy generalisation performance %
	for all three losses with and without regulariser evaluated on central crops. For the losses we use $\ktrain=16$ on CIFAR and $\ktrain=8$ on ImageNet. The NFNet-F0 uses RandAugment as \protect\gls{da} in addition to the horizontal flips and random crops used for the WRN and NF-ResNet. For further results, please see \cref{sec:app:additional_results}.
	}
	\label{tab:results}
\end{table*}

Perhaps a somewhat counter-intuitive result from \cref{fig:results_no_kl} is that the Bayesian-inspired losses lead to worse performance and less invariant predictions as we increase $\ktrain$.\footnote{This observation contradicts results by \citet{Nabarro2021-bv}. While we were able to reproduce their results using their model (a ResNet18 that uses batch normalisation), their observations do not extend to any of our normaliser-free models. For completeness we include a comparison on their model in \cref{app:resnet18}}
We make two arguments to explain this observation. \\
First, we illustrate in an example that the Bayesian-inspired objectives can easily have many non-invariant solutions to simple problems with symmetries compared to \lossstd. 
We consider a simple $2$-dimensional binary classification problem with azimuthal (rotational) symmetry in \cref{fig:toy_example_no_kl} and use a linear $2$-parameter model that is rotationally invariant when its parameters are equal. The loss surface for $\lossstd$ only has a single minimum, which is invariant. However, since $\losslog$ only constrains the \emph{average} logit value, the loss surface for $\losslog$ is also minimised anywhere along a line of non-invariant models, for which the individual logit values vary with the angle (\gls{da}).\\
Second, we note that when we average the losses we make a prediction with every one of the $\ktrain$ augmentations and every one of them has to explain the (same) label, thus encouraging the predictions to be similar. In contrast, for $\losslog$ (or $\lossprob$) we only make a single prediction using the average of the logits (or probabilities) of the individual augmentations. How similar or different the individual values %
are is irrelevant as only their average matters. We speculate that as $\ktrain$ grows, it is sufficient for some of the augmentations to have confident enough predictions that dominate the average, such that there is very little pressure for some of wrong or less confident augmentations to explain the data well, and this pressure might only decrease as $\ktrain$ increases. 

\section{A KL-regularised objective leads to more invariance and better generalisation}
\label{sec:kl_loss}

\begin{figure}[t]
    \centering
    \includestandalone[mode=image]{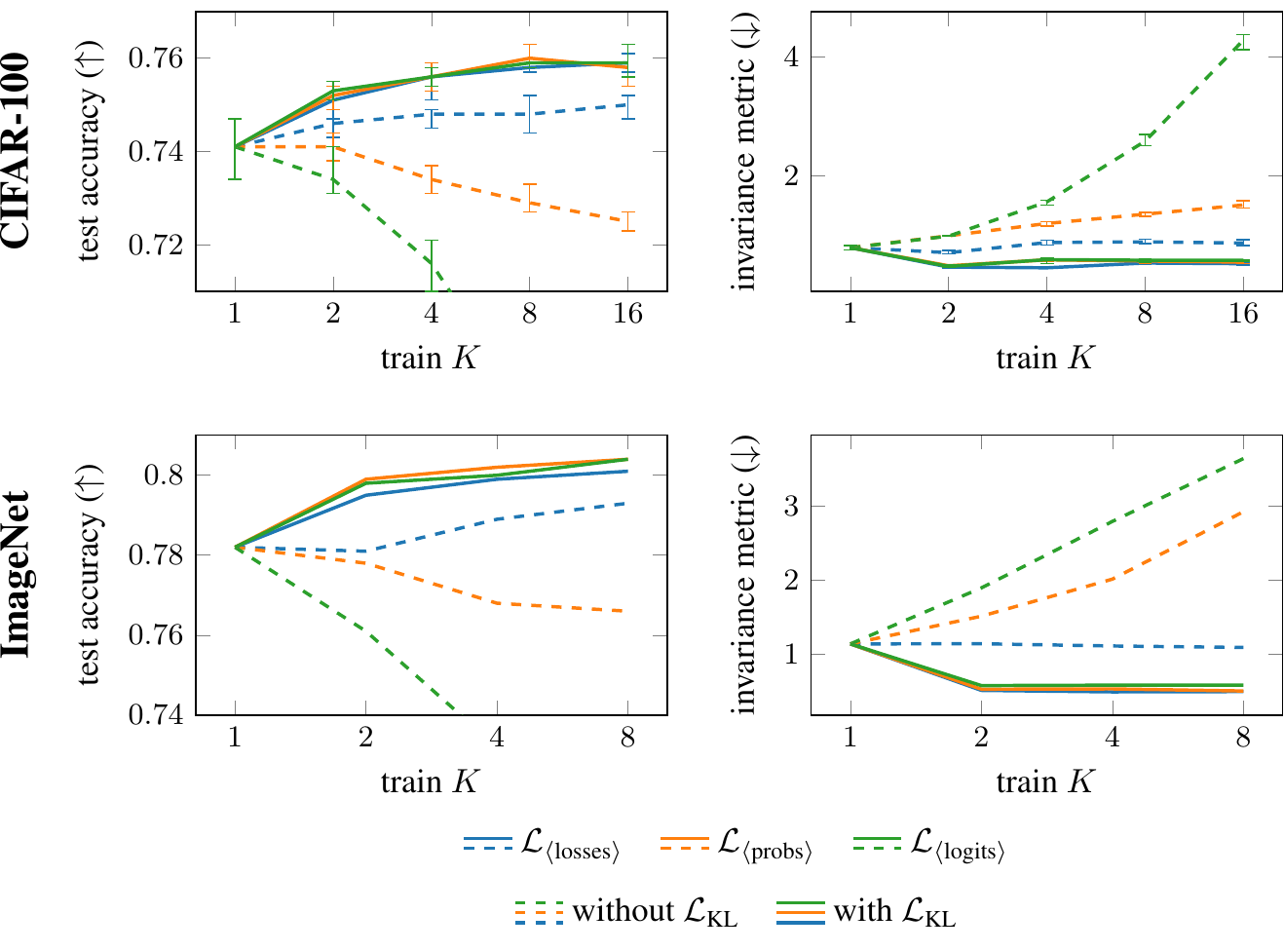}
    \vskip-0.5em
    \caption[]{
    Adding the proposed KL regulariser \losssoft makes the individual predictions more invariant to \gls{da} and further boosts generalisation performance (shown here: top-1 test accuracy on central crops).
    }
    \label{fig:results_kl}
\end{figure}

In \cref{sec:avg_methods} we showed that averaging the losses over \glspl{da} results in better generalisation compared to averaging the logits or probabilities. We hypothesised that this improvement in performance was due to the more invariant individual predictions for \lossstd.
To further investigate this hypothesis, we propose to explicitly regularise the parametric model $\vf_{\vphi}$ to make more similar predictions across individual \glspl{da} of the same input. 
We then study the effect of this invariance regulariser on the generalisation performance of the model.

While the classification losses achieve some degree of invariance, we wish to encourage this objective more strongly and directly. Since the ultimate quantity of interest in supervised learning is the predictive distribution over labels, $p(y\given g(\vf_{\vphi}(\vx))$, a natural choice for such a regulariser is to penalise the KL divergence between predictives for different \glspl{da}. This is in contrast to many self-supervised methods that penalise differences between vectors in an arbitrarily chosen embedding space \citep{Chen2020-simclrv2}.
The KL divergence is attractive since it regularises the entire distribution and not just incorrect predictions. Moreover, it is an information theoretic quantity measured in nats like the original log-likelihood loss (\cref{eq:loss_orig}), which makes it easier to compare their values and reason about the regularisation strength. 

Since the KL divergence is asymmetric, we use its symmetrised version, also referred to as Jeffrey's divergence: 
\begin{align}
    \losssoft(\vx) = \expect{\substack{\vxa,\vxap \sim \pxa}}{\kl{p(y\given\vxa)}{p(y\given\vxap)}}.
    \label{eq:kl_loss}
\end{align}
We add \losssoft as a soft-constraint regulariser to the original objectives \cref{eq:loss_std,eq:loss_avg_probs,eq:loss_avg_logs} with a regularisation strength $\lambda$:
\begin{align}
    \loss_{\left\langle\dots\right\rangle,\text{ regularised}}(\vx, y) = \loss_{\left\langle\dots\right\rangle}(\vx, y) + \lambda \losssoft(\vx).
    \label{eq:full_loss}
\end{align}
We found that $\lambda=1$ works well in practice and fix $\lambda$ to this value. Since the loss and the regulariser are on the same scale, this is a natural choice. We discuss this further in \cref{sec:ablations:regularisation_strength}. We also emphasise that we define the regulariser using the predictive distributions of \emph{individual} augmentations even though \lossprob and \losslog only make a single prediction using \emph{average} probabilities or logits, respectively. %

Similar KL divergence-based regularisers have been explored as objectives in the contrastive and self-supervised learning literature recently; for example, \citet{xie2020unsupervised} use a cross-entropy to regularise predictions on unlabelled data in semi-supervised learning, while \citet{Mitrovic2021-relic} target the predictive distribution of surrogate task-labels in fully unsupervised learning.

In practice we replace the expectation in \cref{eq:kl_loss} by a Monte Carlo estimate $\losssofthat^K$ with size determined by the augmentation multiplicity $K$ used to evaluate the main objective:
\begin{tcolorbox}[enhanced,colback=white,%
    colframe=C1!75!black, attach boxed title to top right={yshift=-\tcboxedtitleheight/2, xshift=-.75cm}, title=Invariance regulariser used as soft-constraint, coltitle=C1!75!black, boxed title style={size=small,colback=white,opacityback=1, opacityframe=0}, size=title, enlarge top initially by=-\tcboxedtitleheight/4, 
    left=-5pt, %
    ]
\begin{equation}
    \losssofthat^K(\vx) = \tfrac{1}{K^2-K} \!\!\sum_{\substack{k, k'=1 \\ k \neq k'}}^K \kl{p(y\given\vxa_k)}{p(y\given\vxa_{k'})}.
    \label{eq:kl_loss_k}
\end{equation}
\end{tcolorbox}
\Cref{eq:kl_loss_k} is an unbiased estimate of \cref{eq:kl_loss} (see \cref{sec:app:loss_comparisons} for the derivation), and we illustrate it for $K=3$ in \cref{fig:method} (c).

\subsection{Experimental evaluation}

To evaluate the effect of the KL regulariser, we add it to the three objectives discussed in \cref{sec:avg_methods} and run otherwise identical experiments on the WideResNet 16-4 with SkipInit on CIFAR-100 and the NF-ResNet-101 on ImageNet. See \cref{sec:app:experimental_details} for more experimental details. 

The qualitative results again agree on both experiments (see \cref{fig:results_kl} and \cref{tab:results}). 
We find that the regularised objectives consistently perform better than their non-regularised counterparts.
Perhaps more interestingly, all three regularised objectives now generalise equally well. Furthermore, we find that the invariance measure now is almost identical for all three regularised objectives as we vary the train augmentation multiplicity. 

These results support our conjecture that the main driver of generalisation performance when using multiple \glspl{da} is the invariance of the \emph{individual} predictions; simply constructing an invariant predictor through averaging non-invariant features, as is done in the Bayesian-inspired losses, is not sufficient . When we account for this by adding our proposed regulariser, which further encourages invariance on the level of individual predictions, all objectives improve and now show similar performance. 

\section{Ablations}
\label{sec:ablations}

In this section, we provide additional experiments that support our main findings. In particular, we show that our results also hold for a much wider range of data augmentations, and when changing the capacity of our models. We also study the effect of varying the strength $\lambda$ of the KL regulariser.
In addition we include extended experimental results in \cref{sec:app:additional_results}.

\subsection{Using a larger set of data augmentations}
\label{subsec:more_augs}
\begin{figure}[t]
    \centering
    \includestandalone[mode=image]{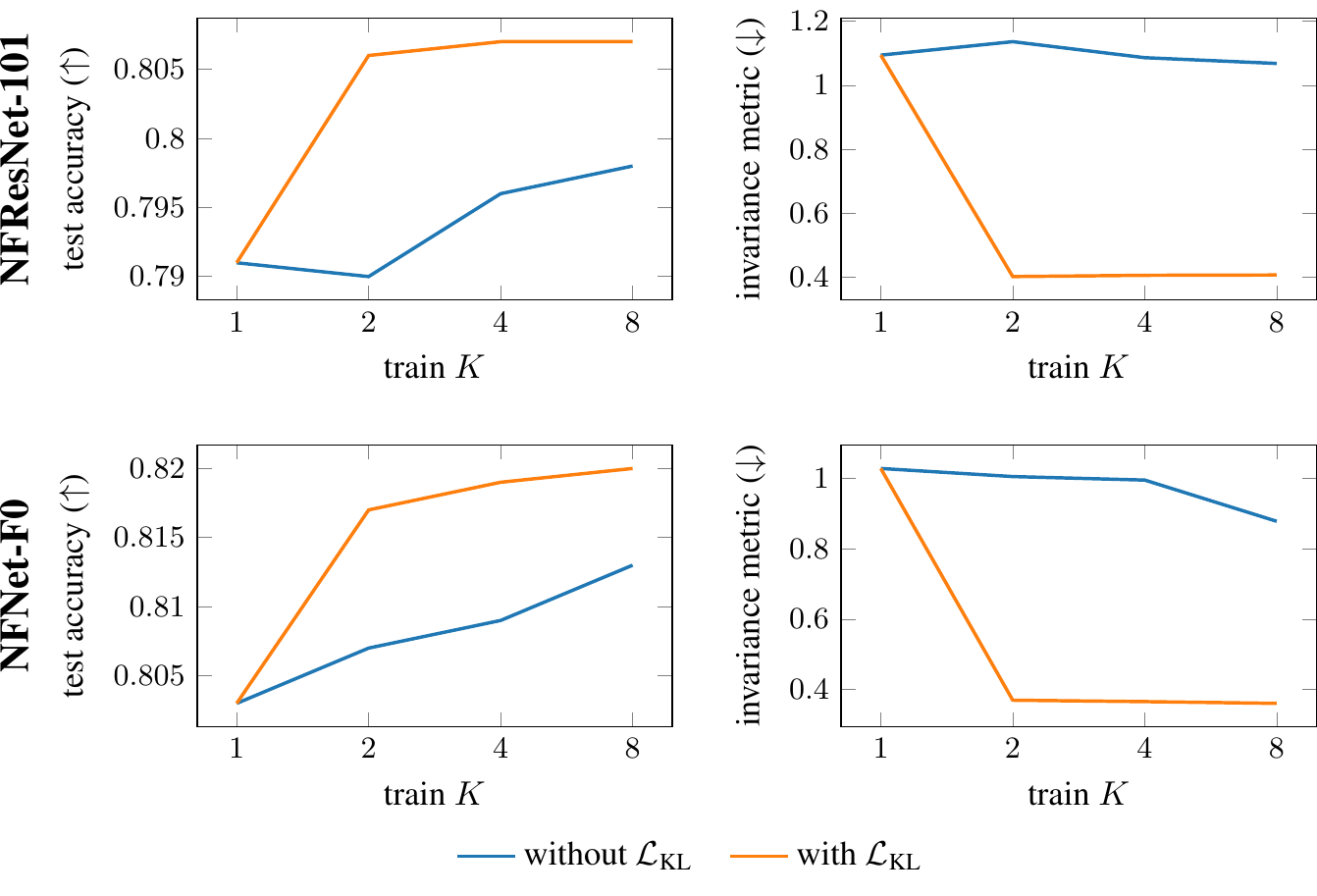}
    \vskip-0.5em
    \caption[]{Networks trained on ImageNet with \lossstd and stronger data augmentations (RandAugment).
    }
    \label{fig:resnet_randaugment}
\end{figure}

In all previous experiments we only used horizontal flips and random crops as \glspl{da}.
In this section, we show that our main findings also apply more generally when using stronger \glspl{da} and for more modern image classification architectures, namely NFNets \citep{brock2021high}. As stronger \glspl{da}, we consider RandAugment \citep{randAugment} in addition to horizontal flips and random crops, which use a combination of 16 augmentations such as colour and brightness changes or image rotations, shears, and distortions. We set the magnitude of RandAugment to $5$ following the setting used for the NFNet-F0 in \citet{brock2021high}. 

We first train the NF-ResNet-101 on ImageNet with the same experimental setup as before but adding RandAugment to the set of \glspl{da}. We then also train an NFNet-F0 \citep{brock2021high} on ImageNet; the NFNet models are highly expressive, and therefore prone to overfit, and rely on strong \glspl{da} to achieve good generalisation performance \citep{brock2021high}. We exclude any \glspl{da} that use mixing between different inputs, such as CutMix \citep{yun2019cutmix} and Mixup \citep{zhang2017mixup} such that the model only uses RandAugment, horizontal flips and random crops as \gls{da}. 
Due to computational constraints, we only run experiments with the $\lossstd$ objective and its regularised version for different augmentation multiplicities.
We also modify the training procedure to make the setting more comparable to our other experiments; see \cref{sec:app:experimental_details} for details. 
While these modifications slightly reduce performance of the NFNet-F0, the baseline model ($\ktrain=1$) still achieves $80.3\%$ top-1 accuracy on central crops compared to $83.6\%$ as reported by \citet{brock2021high}.

For both experiments, the results shown in \cref{fig:resnet_randaugment} qualitatively agree with those discussed in \cref{sec:kl_loss}. In particular, we see that the KL regulariser improves performance over the non-regularised counterpart, and again results in significantly more invariant predictions.

\begin{figure}
    \centering
    \includestandalone[mode=image]{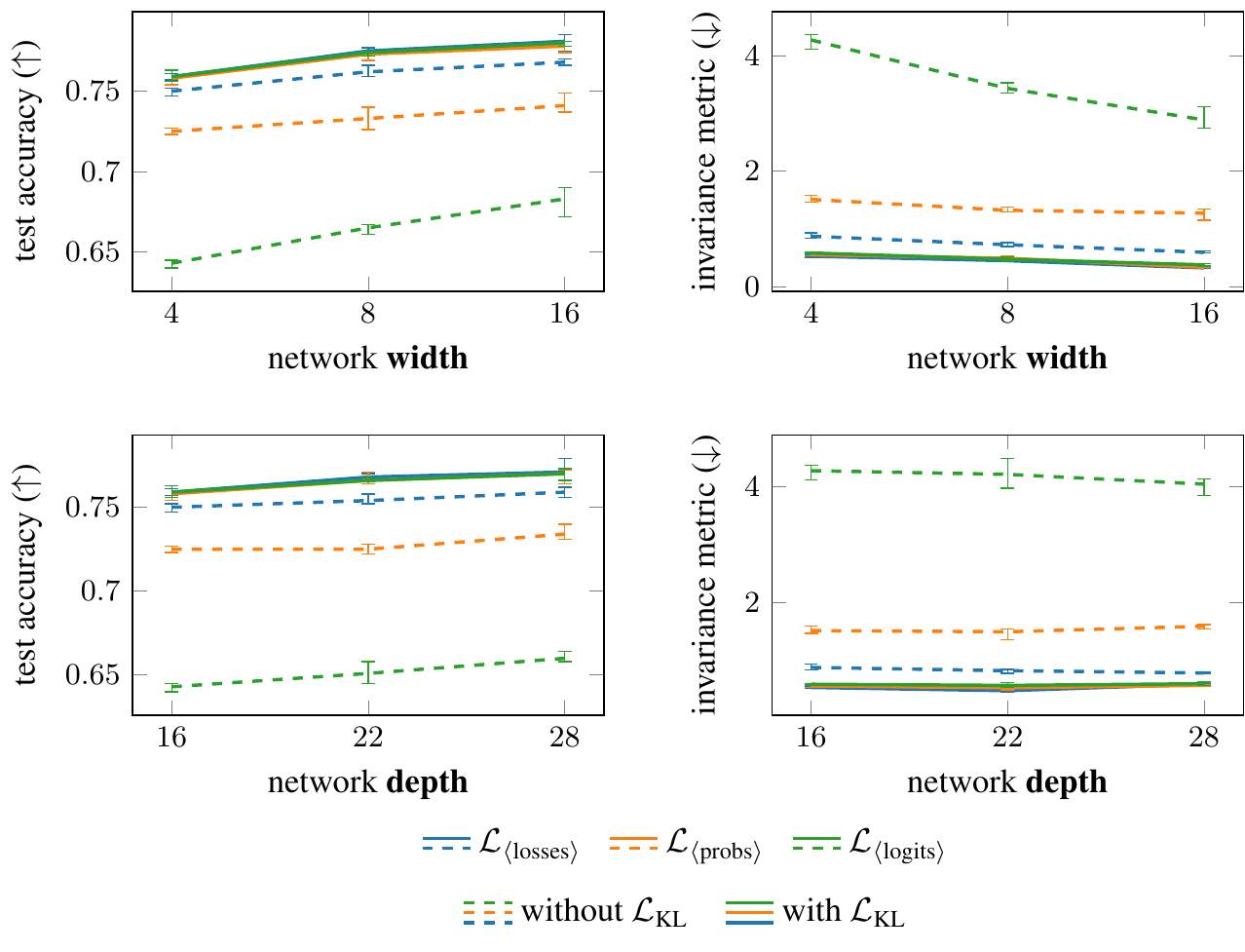}
    \vskip-0.5em
    \caption{WideResNets with varying width factors and depths on CIFAR-100. We train with $\ktrain=16$ and evaluate on central crops.
    }
    \label{fig:cifar_wrn_wXdY}
\end{figure}

\subsection{Effect of changing model capacity}

In this section, we investigate the effectiveness of the KL regulariser as the capacity of the model changes. 
For the WideResNet on CIFAR-100 we run additional experiments with varying width and depth. At depth $16$ we consider width factors $4$ (default), $8$, and $16$; at width factor $4$ we consider depths $16$ (default), $22$, and $28$.
For each model we tune the optimal learning rate and optimal epoch budget. 
From the results in \cref{fig:cifar_wrn_wXdY} we can observe that as model capacity increases (both for width and depth), so does the test accuracy of the baselines as expected. 
More importantly, the models trained with our regulariser improve almost in parallel as well, with the generalisation gap between the two remaining identical. Furthermore, in each case the regulariser also equalises the performance between the three methods. Interestingly, while the invariance measure gets better for wider networks, it stays roughly the same as the network gets deeper.

\subsection{Sweeping the strength of the KL regulariser}
\label{sec:ablations:regularisation_strength}

\begin{figure}[t]
    \centering
    \includestandalone[mode=image]{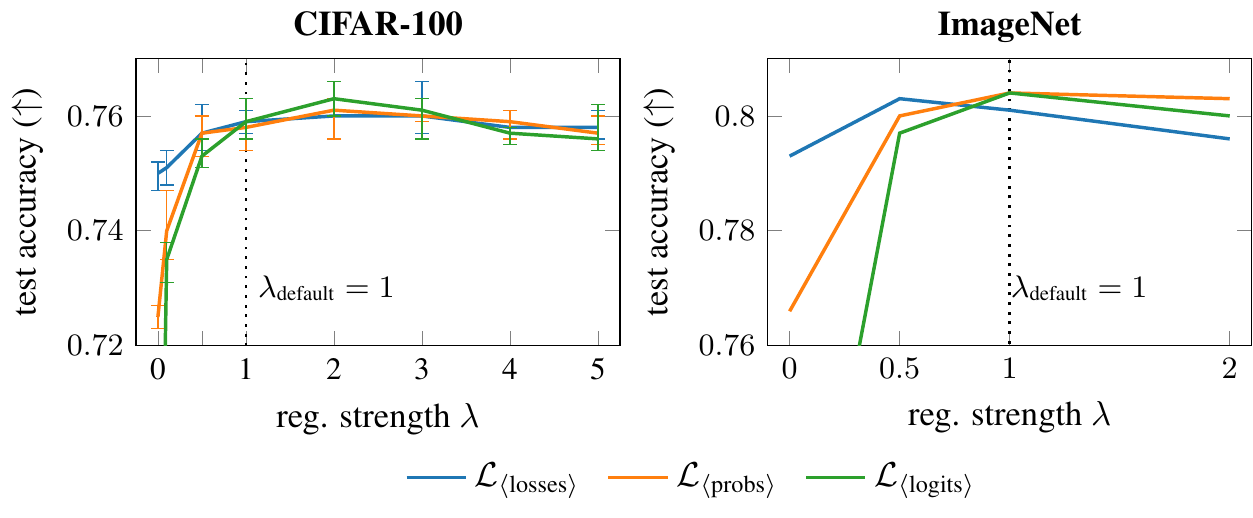}
    \vskip-0.5em
    \caption{Sensitivity of results to the choice of regularisation strength $\lambda$. NF-ResNet-101 with $\ktrain=8$ on Imagenet, and WideResNet 16-4 with $\ktrain=16$ on CIFAR-100, both evaluated on central crops.}
    \label{fig:ablation_vary_lambda}
\end{figure}

In all our previous experiments, we set the strength of the KL regulariser to $\lambda = 1$. As mentioned in \cref{sec:kl_loss}, one important reason why the KL divergence might be well suited in practice as a regulariser is that it is measured in the same units as the log-likelihood, \emph{nats per image}, and hence is on the same scale as the original objective.
This suggests that %
$\lambda \approx 1$ should be a good value for $\lambda$ as it provides equal balance between the original objective and the regulariser. 

Here, we provide a comparison when sweeping over the value of $\lambda \in [0, 2 \text{ or } 5]$ for the WideResNet on CIFAR-100 and the NF-ResNet-101 on ImageNet, respectively.
The corresponding test accuracies using central-crop evaluation are shown in \cref{fig:ablation_vary_lambda}. We find that the generalisation performance is relatively insensitive to the choice of $\lambda$ for large enough values, and that $\lambda=1$ consistently yields among the best performance. $\lambda=0$ corresponds to training without the regulariser and performs markedly worse. Very large values of $\lambda$ also lead to degraded accuracies and likely correspond to over-regularization. These results suggest that the regulariser weight $\lambda=1$ is close to optimal, though minor gains are possible through further tuning of $\lambda$.

\section{Related Work}
\label{sec:related_work}
\paragraph{Data-augmentation and invariance in neural networks.}

The incorporation of invariances is a common inductive bias in machine learning and different approaches have been developed over the years.
First, \gls{da} methods that enlarge the training set and implicitly enforce a model's predictions to be correct for a larger portion of the input space \citep{Niyogi1998-data-augmentation,Beymer1995-data-augmentation}. 
The augmentations are dataset-dependent and hand-tuned \citep{randAugment, yun2019cutmix, cutout} or sometimes even learned \citep{Wilk2018-learning-invariances,autoaugment}.
Second, methods that explicitly constrain the intermediate network or its output to be invariant or equivariant to certain transformations. For example, 2D convolutions \citep{LeCun1989-xf,Lecun1998-oe} are equivariant to translations while generalised convolutions \citep{Cohen2016-equivariant-convolution} are equivariant to more general group transformations. However it is difficult to hard-code more complex equivariances/invariances.
\citet{Wilk2018-learning-invariances,Raj2017-on} recently proposed to construct an invariant covariance function for kernel methods by integrating a non-invariant kernel over the augmentation distribution. %
\citet{Nabarro2021-bv} use this approach to define invariant losses for neural networks by averaging the logits or probabilities as discussed in \cref{sec:avg_methods}.
Third, regularisation methods that do not place hard constraints on the classifier function or its outputs but instead add an additional loss as a soft-constraint that encourages the desired behaviour. Examples from 
the self-supervised literature (see next paragraph) and our proposed KL soft-constraint fall into this category.
\citet{Bouchacourt2021-fb-invariance-grounding} recently investigated the implicit effects of \glspl{da} on how invariant models are. 
Though, they focus on sampling a single augmentation per image, whereas we consider training with larger augmentation multiplicities ($K > 1$).

We note that concurrently with this work, other works have explored explicit \gls{da} regularisers for empirical risk minimisation. \citet{Yang2022_data_augmentation_consistency} show that explicitly regularising the features of different augmentations of the same input to be similar can
achieve better generalisation than training with empirical risk minimisation on an augmented data set.
\citet{Huang2021-dair} propose and study an explicit regulariser in this vein showing improvements over standard training, though they only consider pairs of augmentations for each update.

\paragraph{Self-supervised and contrastive methods} are commonly used to learn  visual representations from unlabelled data. 
Simply speaking, they construct a surrogate ``self-'' supervised learning problem and use it to train a neural network feature extractor that is subsequently used in other downstream tasks.
For example, \citet{Doersch2015-unsupervised-context,Noroozi2016-jigsaw,Gidaris2018-unsupervised-rotation} propose hand-crafted tasks such as solving a jigsaw. 
More recent approaches use \glspl{da} to construct surrogate instance discrimination tasks and directly maximise a similarity measure between projected features for different augmentations of the same image \citep{Grill2020-byol} or solve corresponding clustering problems \citep{Caron2020_swav}. 
Contrastive methods additionally maximise the discrepancy between augmentations of different images \citep{Chen2020-simclr,Chen2020-simclrv2}. 
In addition to a contrastive loss, \citet{Mitrovic2021-relic} use a KL regulariser similar to ours; however, where we use the predictive for the true label, they regularise probabilities of a surrogate task. 
Similar regularisers have also been considered for unlabelled data in semi-supervised settings; for example, \citet{Sajjadi2016-semisupervised} minimise the L2-distance between features, and \citet{xie2020unsupervised} propose a cross-entropy-regulariser. Most methods cited above (with the exception of \citet{Mitrovic2021-relic,Sajjadi2016-semisupervised}) only consider pairs of augmentations, whereas we use larger augmentation multiplicities that additionally improve performance.

\section{Conclusion}
In this paper we investigated implicit and explicit regularisation with data augmentation in supervised learning. We discussed two approaches that both use multiple data augmentations per input but differ in how and at what level they encourage or enforce invariance to data augmentation in the predictor: (i) by averaging the losses of individual augmentations to encourage invariance on the level of the network outputs; or (ii) by averaging the logits or probabilities to make the whole predictor invariant by construction, though network outputs on individual augmentations are not necessarily invariant. 
We found empirically that the former approach generalises better and that its outputs are more similar across different augmentations of the same image.
Motivated by this, we introduced a KL regulariser which explicitly encourages this similarity of the network outputs.
Through extensive experiments on CIFAR-100 and ImageNet with multiple large-scale models, we showed that the proposed regulariser improves generalisation performance for all methods and largely equalises performance of the considered approaches. 
Our results confirm that encouraging invariance on the level of the individual predictions drives the improvements in generalisation performance when using multiple augmentations per image.

\clearpage

\subsection*{Acknowledgements}
We thank Andriy Mnih for helpful discussions and feedback on our analysis, Mark van der Wilk for discussions on Bayesian data augmentation, as well as Razvan Pascanu and Yee Whye Teh for feedback on the manuscript. 

\printbibliography

\clearpage
\appendix
\onecolumn
\section{Further analysis of the objectives and the KL regulariser}
\label{sec:loss_comparisons}
\label{sec:app:loss_comparisons}

In this section we further analyse and relate the three objectives \cref{eq:loss_std,eq:loss_avg_probs,eq:loss_avg_logs} as well as the KL regulariser and discuss their finite sample estimators.

\subsection{The three objectives using multiple data augmentations}
First, we analyse \lossstd, \lossprob, and \losslog in more detail. For convenience, we reproduce their defenitions here:
\begin{align}
    \lossavglosses(\vx, y) & = \expect{\vxa \sim p(\vxa \given \vx)}{\varL_{\vphi}(\vxa, y)} \tag{\ref{eq:loss_std}} \\
    \lossavgprobs(\vx, y) &= -\log \expect{\vxa \sim p(\vxa \given \vx)}{p(y \given g(\fphi(\vxa)))}, \tag{\ref{eq:loss_avg_probs}}\\
    \lossavglogs(\vx, y) &= -\log p(y \given g(\expect{\vxa \sim p(\vxa \given \vx)}{\fphi(\vxa)})) \tag{\ref{eq:loss_avg_logs}}
\end{align}
While $\lossstd$ is not a valid likelihood \citep{Wenzel2020-cold-posterior,Nabarro2021-bv}, we can show that it constitutes a lower bound to both $\losslog$ and $\lossprob$:
Using their definitions from \cref{eq:loss_std,eq:loss_avg_probs,eq:loss_avg_logs} and the fact that the $\log$ as well as the $\log\softmax$ are concave functions, it immediately follows from Jensen's inequality that:
\begin{equation}
    \begin{aligned}
        \losslog \geq \lossstd \quad\text{and} \quad \lossprob \geq \lossstd
    \end{aligned}
    \label{eq:loss_bounds}
\end{equation}
Here, we have assumed that the inverse link function $g$ corresponds to the $\softmax$. The bounds in \cref{eq:loss_bounds} more generally hold if the composition $\log\circ g$ is concave.\\
Note that despite of this result, there is no strict relationship between $\losslog$ and $\lossprob$, as the $\softmax$ function by itself is neither convex nor concave.

\subsection{Finite sample objectives for the losses}

In practice, we always have to approximate the marginalisation of the augmentation distribution in \cref{eq:loss_std,eq:loss_avg_probs,eq:loss_avg_logs} by a finite average over $K$ Monte Carlo samples. This gives rise to the finite sample estimators discussed in the main paper:
\begin{align}
    \lossstdhat^K(\vx, y) &= \tfrac{1}{K}\textstyle\sum_{k=1}^K \varL\left(g(\fphi(\vxa_{k})), y\right) \tag{\ref{eq:avg_losses_k}}\\
    \lossprobhat^K(\vx, y) &= \varL\left(\tfrac{1}{K}\textstyle\sum_{k=1}^K g(\fphi(\vxa_{k})), y\right) \tag{\ref{eq:avg_probs_k}}\\
    \lossloghat^K(\vx, y) &= \varL\left(g\left(\tfrac{1}{K}\textstyle\sum_{k=1}^K\fphi(\vxa_{k})\right), y\right) \tag{\ref{eq:avg_logs_k}}
\end{align}
The natural question is, how well these estimators approximate the original losses and whether they are unbiased.

To answer this question, we define $p_K(\vxa_{1:K} \given \vx)$ as the joint distribution over $K$ independent samples from the augmentation distribution, that is
\begin{equation}
    p_K(\vxa_{1:K} \given \vx) = \prod_{k=1}^K p(\vxa_k \given x).
\end{equation}
Then, it is easy to see that $\lossstdhat^K$ is an unbiased estimator of $\lossstd$ because we can exchange summation and expectation. This is not possible for $\lossloghat^K$ and $\lossprobhat^K$ due to the non-linearity of the $\log$ and the inverse link function $g$.
However, \citet{Nabarro2021-bv} proved that they form lower bounds for any value of $K$:
\begin{equation}
\begin{aligned}
    \lossprob &\ge \expect{p_K}{\lossprobhat^K} \ge \expect{p_1}{\lossprobhat^1} = \lossstd, \\
    \losslog &\ge \expect{p_K}{\lossloghat^K} \ge \expect{p_1}{\lossloghat^1} = \lossstd.
\end{aligned}
\label{eq:loss_bounds_sampling}
\end{equation}
Moreover, they showed that the bounds get tighter as $K$ increases, that is, the expectations $\expect{p_K}{\lossprobhat^K}$ and $\expect{p_K}{\lossloghat^K}$ are increasing functions in $K$.

Similar to \cref{eq:loss_bounds} we can again use Jensen's inequality to show that for any fixed finite set of augmentations $\vxa_{1:K}$ of size $K$ taken from $p_K$ that $\lossstdhat^K$ is a lower bound of the other two estimators:
\begin{equation}
\begin{aligned}
    \lossloghat^K \geq \lossstdhat^K \quad\text{and} \quad \lossprobhat^K \geq \lossstdhat^K.
\end{aligned}
\label{eq:loss_bounds_sampling}
\end{equation}
Again, there is no strict relation between $\lossprobhat^K$ and $\lossloghat^K$.

\subsection{Finite sample objective for soft-constraint}

For completeness, we here show that the finite sample estimator of the regulariser $\losssofthat^K(\vx)$, \cref{eq:kl_loss_k} is an unbiased estimator of full regulariser \cref{eq:kl_loss}. 

In fact for any function $f(\vxa, \vxa')$ it is true that:
\begin{equation}
\begin{aligned}
    &\expect{\vxa_{1:K} \sim p_K(\vxa_{1:K}\given\vx)}{\tfrac{1}{K^2-K} \!\!\sum_{\substack{k, k'=1 \\ k \neq k'}}^K f(\vxa_k, \vxa_{k'})}= \\ 
    & \quad= \tfrac{1}{K^2-K} \!\!\sum_{\substack{k, k'=1 \\ k \neq k'}}^K \expect{\vxa_{1:K} \sim p_K(\vxa_{1:K}\given\vx)}{f(\vxa_k, \vxa_{k'})} \\
    & \quad= \expect{\vxa_{1:K} \sim p_K(\vxa_{1:K}\given\vx)}{f(\vxa_1, \vxa_2)} \\ 
    & \quad = \expect{\substack{\vxa, \vxa' \sim \pxa}}{f(\vxa, \vxa')}.
\end{aligned}
\end{equation}
Replacing $f$ by the KL-divergence yields the desired result.

\section{Software}
\label{sec:app:software}
We use \texttt{numpy} \citep{Harris2020-numpy} and build our networks in \texttt{jax} \citep{Bradbury2018-jax} using \texttt{haiku} \citep{Hennigan2020-haiku}.

\section{Experimental details}
\label{sec:app:experimental_details}

We describe additional details for our experiments in this section.

\subsection{Network architectures and training procedure}

In this section, we describe more details about the network architectures used and the training and hyperparameter selection procedures used in our experiments.

\subsubsection{Wide ResNets}
Since we focus on models that do not have batch normalisation, we use a WideResNet with the SkipInit initialisation scheme from \citet{de2020batch} with a depth of $16$ and a width factor of $4$, referred to as WRN-16-4.
Additionally, for the ablation study in \cref{sec:ablations} we grow either the width or the depth of the model. 
For growing widths we consider WRN-16-8 and WRN-16-16 models and for growing depth WRN-22-4 and WRN-28-4 models. 

For training we use a total batch size of $64$ and run experiments with augmentation multiplicity of $K\in \{1, 2, 4,8, 16\}$, meaning that for $K=16$ a single batch will contain $4$ unique images from $\data$ and to each one we apply $16$ random augmentations sampled independently. 
For each model and objective we sweep and find the optimal epoch budget $E$ from the set $\{64, 96, 128, 192, 256\}$.
All models have a weight decay of $0.0005$.
For optimisation we use stochastic gradient descent with momentum.
The momentum value is $0.9$ and the base learning rate is defined as $B_{\text{unique}} 2^{\alpha}$, where $B_{\text{unique}}$ is the number of unique images per batch and $\alpha$ is a hyper parameter.
We keep the base learning rate constant for the first $\frac{E}{2}$ epochs, and then for the remainder of training, we reduce the learning rate by a factor of $2$ every $\frac{E}{20}$ epochs. 
This learning rate schedule achieves similar performance to other decay schedules used for these models while having a single hyperparameter $\alpha$ which we tune from the values in $\{-8, -9, -10, -11, -12, -13\}$. 

\subsubsection{NF-ResNets}

For training we use a total batch size of $1024$ and run experiments with augmentation multiplicity of $K\in \{1, 2, 4, 8\}$, meaning that for $K=8$ a single batch will contain $128$ unique images from $\data$ and to each one we apply $8$ random augmentations sampled independently. 
For each model and objective we sweep and find the optimal epoch budget $E$ from the set $\{64, 128, 256, 512, 1024\}$.
We use weight decay of $0.0005$ and label smoothing with strength $0.1$.
Additionally, as is common we use dropout with rate $0.25$ and a stochastic depth drop rate of $0.1$ \citep{brock2021high}.
For optimisation we use stochastic gradient descent with momentum.
The momentum value is $0.9$ and for the learning rate we use a cosine decay schedule defined as $2^{\alpha - 1} \left(1 + \cos (\pi s )\right)$, where $s$ is the current step counter divided by the total number of steps the optimiser will perform.
The only hyperparameter $\alpha$ is tuned from the values in $\{-1, -2, -3, -4\}$. 

\subsubsection{NFNet-F0}
To be consistent with the rest of our experiments in this paper, we use SGD with momentum (with momentum coefficient 0.9) as our optimiser, and do not use Adaptive Gradient Clipping (AGC) as in \citep{brock2021high}. To partially counteract for this, we use a batch size of 1024 (instead of a batch size of 4096). We use RandAugment with a magnitude of 5. We also remove any \glspl{da} that use mixing between different inputs, such as CutMix \citep{yun2019cutmix} and Mixup \citep{zhang2017mixup}, as used with the original NFNet models \citep{brock2021high}. We also used an image size of 224 for both training and evaluation, consistent with the rest of the ImageNet experiments in our paper, while \citet{brock2021high} used an image size of 192 for training and an image size of 256 for testing on the NFNet-F0. While all these modifications slightly reduce the performance of the NFNet-F0, the baseline model still achieves a relatively high $80.3\%$ top-1 accuracy, compared to $83.6\%$ as reported by \citet{brock2021high}.

Similar to \citet{brock2021high}, we use a warmup learning rate schedule, where the learning rate is increased from 0 to the specified value over the first 5 epochs. We also use a cosine annealing decay schedule for the learning rate for the rest of training, as described in the previous section, which decays the learning rate down to 0. We use a weight decay of $0.00002$, label smoothing with a parameter of $0.1$, dropout on the classification layer with drop rate $0.2$, and a stochastic depth drop rate of $0.25$, all following \citep{brock2021high}.

Due to computational constraints, we employ a slightly smaller sweep to find the optimal learning rate: we sweep the values of $\{0.2, 0.4, 0.8\}$. To find the optimal epoch budget for each augmentation multiplicity, we sweep the epoch budget with the values $\{45, 90, 180, 360\}$. Note that the NFNet-F0 in \citet{brock2021high} were trained for 360 epochs, and as augmentation multiplicity increases, we would expect the optimal epoch budget to decrease, as evidenced by \citet{Fort2021-yq}.

\subsection{Evaluation}
\label{sec:app:details_evaluation}

We use two main methods for evaluation: (1) central crops, and (2) test-time augmentations (TTA).

\paragraph{Central crops.} We use a single central crop from the input image. This is the standard approach to evaluate performance in supervised learning tasks \citep{krizhevsky2012imagenet}.

\paragraph{Test-time augmentations (TTA).} For \gls{tta}, we use the same augmentation pipeline that is used during training and compute the probabilities for each of the $\ktest$ augmentations separately. We then make an ensemble prediction where we average the probabilities, similar to how \lossprob is defined:
\begin{align}
    y_\text{pred}\given \vx & = \arg\max_y \tfrac{1}{\ktest}\sum_{k=1}^{\ktest} p(y\given g(\vf_{\vphi}(\vxa_k)))
    \label{eq:app:tta_avgprobs}
\end{align}
where $\vxa_{1:\ktest}$ are augmentations sampled i.i.d. from the augmentation distribution $\pxa$. This is the standard approach for \gls{tta} to boost performance over central crops \citep{krizhevsky2012imagenet, simonyan2014very, szegedy2015going}. Though, more sophisticated method exist, for which the weighting or sampling distribution of different augmentation functions are learned \citep{Shanmugam2020-tta} or even predicted from the input image \citep{learned_tta}.

While standard \gls{tta} ensembles the probabilities of different augmentations of the same input, \cref{eq:app:tta_avgprobs}, we could also ensemble their logits instead, similarly to how \losslog is defined:
\begin{align}
    y_\text{pred}\given \vx & = \arg\max_y p\left(y\given g\left(\tfrac{1}{\ktest}\sum_{k=1}^{\ktest}\vf_{\vphi}(\vxa_k)\right)\right)
    \label{eq:app:tta_avglogits}
\end{align}
In the main paper we follow the standard definition of \gls{tta} of ensembling the probabilities; however, for completeness, we also include \gls{tta} where we average the logits in some of the results in this appendix.

For the ablations in \cref{subsec:more_augs} where we also included RandAugment \citep{randAugment} in the set of \glspl{da}, we also include RandAugment in the TTA. Note that RandAugment is an extremely strong \gls{da} technique, and therefore might not reliably improve accuracy performance when employed at test-time, in contrast to weaker \gls{da} methods like horizontal flips or random crops.

\subsection[]{Details on the illustrative example in \protect\cref{fig:toy_example_no_kl}}

For this example we define the true distribution data distribution as:
\begin{align}
    p(\vx) &= \mathcal{N}\left( \begin{bmatrix}1\\1\end{bmatrix}, \begin{bmatrix}2 & 1.5 \\ 1.5 & 2 \end{bmatrix} \right),\\
    p(y &= 1 \given \vx) = \frac{1}{1 + \exp\left(- 0.4 * ||x||_2^2 + 1.5 \right)}.
\end{align}
In \cref{fig:toy_example_no_kl} (left) are shown 50 random datapoints sampled independently. 
The parametric model considered in this case is just a linear classifier, where we assume we know the correct bias term:
\begin{align}
    \fphi(\vx) = w_1 x_1^2 + w_2 x_2^2 - 1.5.
\end{align}
The plots on the middle and right hand side show the loss surface with respect to $w_1$ and $w_2$ in the limit of infinite data.

\section{Comparison on a ResNet-18 with batch normalisation}
\label{app:resnet18}
In this section we briefly discuss results on the ResNet-18 architecture used by \citet{Nabarro2021-bv}. In contrast to our models discussed in the main paper, this ResNet-18 uses batch normalisation.
The performance of models using batch normalisation typically depends strongly on the "ghost batch size", i.e., the number of examples the batch statistics are estimated over \citep{hoffer2017train}. Since the optimal value of this ghost batch size would depend on the augmentation multiplicity, in the main paper we follow \citet{Fort2021-yq} and chose to focus on highly performant models that do not use batch normalisation.

However, for completeness we also reproduce the results of \citet{Nabarro2021-bv} in this section using the ResNet18 with batch normalisation. This is particularly interesting since some of our observations in the main paper on models without batch normalisation contradict the results of \citet{Nabarro2021-bv}.

For training the ResNet18 model, we use a total batch size of $528$ and run experiments with augmentation multiplicity of $\ktrain\in \{1, 2, 4, 8, 16\}$, meaning that for $\ktrain=16$ a single batch will contain $33$ unique images from $\data$ and to each one we apply $16$ random augmentations sampled independently. Note that for this study to be fully rigorous, one should also sweep over the ghost batch size; however, due to computational constraints and to be able to reproduce the results of \citet{Nabarro2021-bv}, we use the same experimental setup as \citet{Nabarro2021-bv} and estimate the batch statistics over all examples in the minibatch. 
We sweep and find the optimal epochs number $E$ from the set $\{272, 288, 304, 320, 336, 352\}$.
We use a weight decay of $0.0005$ and we do not use any dropout or label smoothing.
For optimisation we use stochastic gradient descent with momentum.
The momentum value is $0.9$ and for the learning we use the same cosine decay schedule as in the NF-ResNet experiments, but we also scale the base learning rate by the number of unique images in a batch, similar to the experiments with the WideResNets.
We sweep the hyperparameter $\alpha$ from the values in $\{-9, -9.5, -10.0, -10.5, -11.0\}$. 

\begin{figure*}
    \centering
    \includestandalone[mode=image]{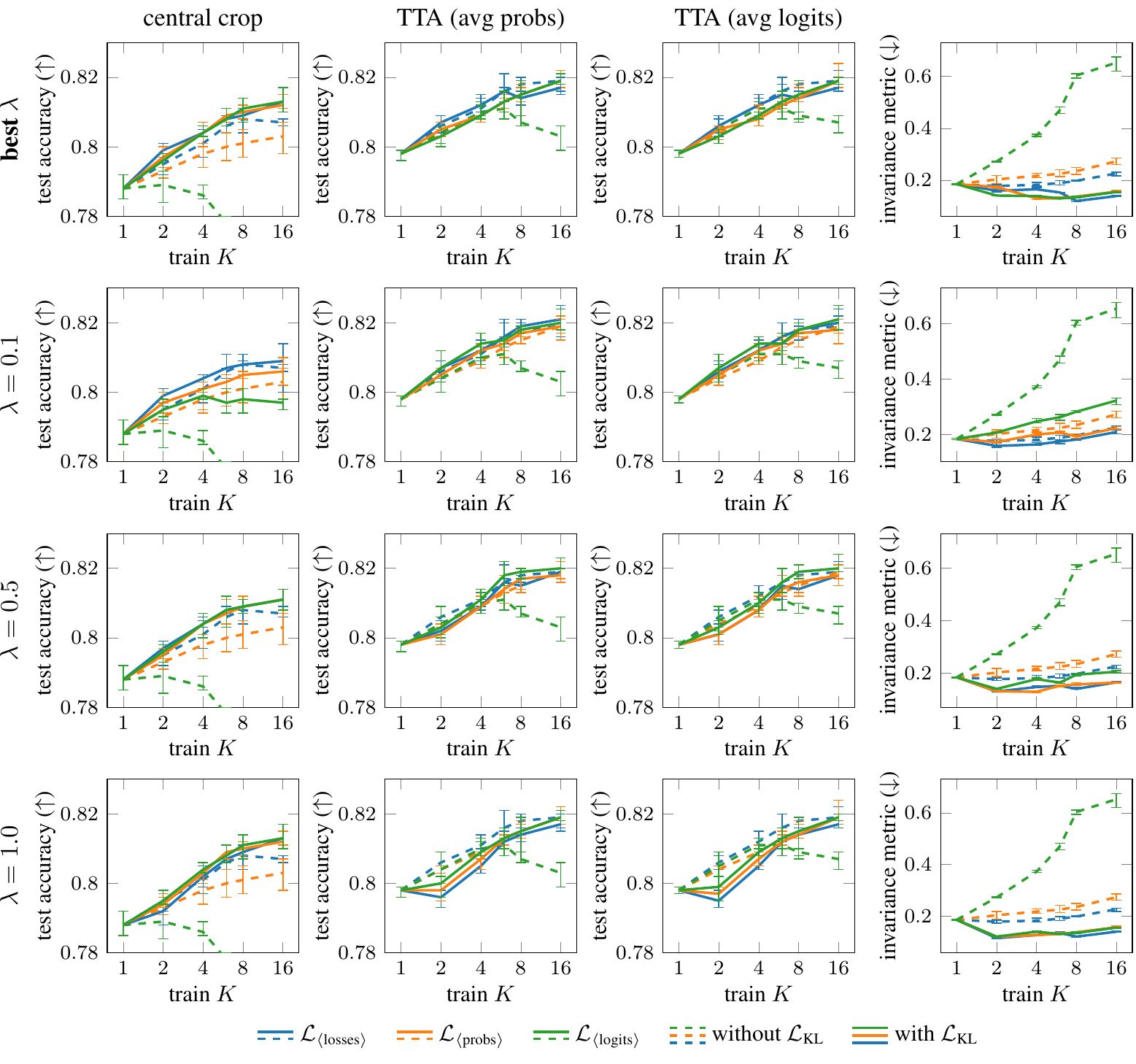}
    \caption{Results on CIFAR-100 with a ResNet18 that uses batch normalisation as reference to compare to results by \citet{Nabarro2021-bv}.}
    \label{fig:app:resnet18}
\end{figure*}

The results are shown in  \cref{fig:app:resnet18}. We find that in contrast to the range of architectures studied in the main paper, we can qualitatively reproduce the results by \citet{Nabarro2021-bv} for \lossprob and \losslog on this architecture. In particular, they reported an \emph{increase} of generalisation performance as the train augmentation multiplicity $\ktrain$ increased for \lossprob as well as on part of the $\ktrain$-range for \losslog. We observe this behaviour (dashed orange and green lines) on central crops as well as when using TTA.\\
Our experiments also show that -- as in all our other experiments -- \lossstd yields better generalisation performance than both \lossprob and \losslog. However, this method has not been considered by \citet{Nabarro2021-bv}.
Moreover, our other qualitative results from the main paper also apply in this setting: (i) without KL regularisation, the \lossstd leads to more invariant predictives, though the difference is smaller than for our other experiments; (ii) with KL regularisation, the performance of all objectives is similar and (at least on central crops) always better than without the regulariser. 

\section{Additional experimental results}
\label{sec:app:additional_results}

Here, we present additional experimental results on CIFAR-100 and ImageNet.

\subsection{Influence of augmentation multiplicity at test time, $\ktest$}
\begin{figure*}
    \centering
    \includestandalone[mode=image]{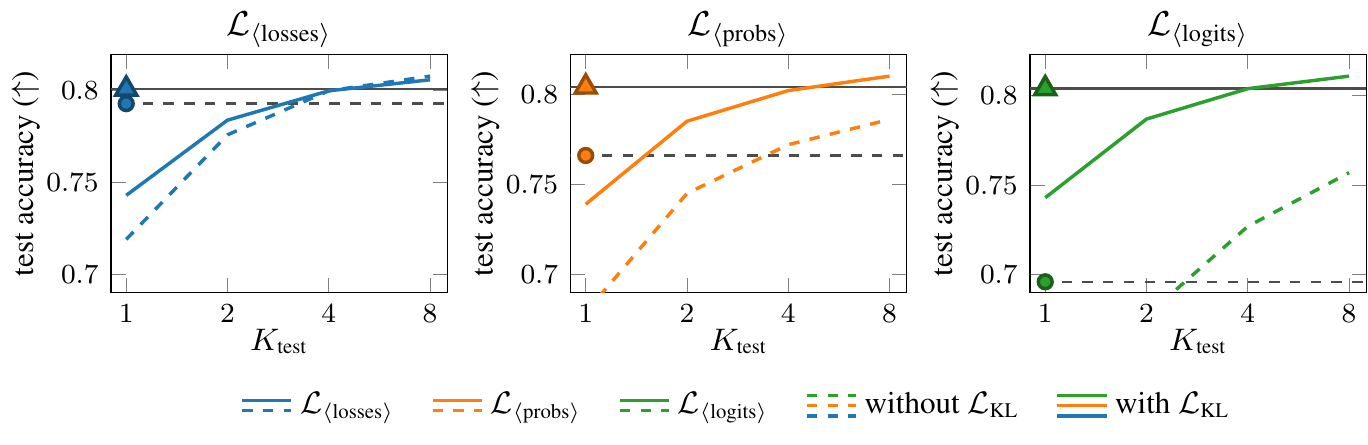}
    \caption{Effect of the augmentation multiplicity at test time $\ktest$ on performance. We consider ImageNet with an NFResNet101 and use training augmentation multiplicity $\ktrain=8$. Triangles and corresponding solid black lines denote central crop performance \emph{with} KL regulariser; circles and corresponding dashed lines denote central crop performance \emph{without} KL regulariser.}
    \label{fig:app:sweep_ktest}
\end{figure*}

Whenever we report test accuracies using test-time augmentations, we show results for the largest value of $\ktest$, which typically corresponds to the largest value of $\ktrain$ ($8$ for ImageNet and $16$ on CIFAR-100). Increasing $\ktest$ always improved performance, and in \cref{fig:app:sweep_ktest} we show a sweep over this value. 

We note that we generally require around at least $\ktest=4$ on most experiments to achieve the same performance as central crops. A value of $\ktest=1$ corresponds to a single augmentation, which is different from using the central crop image.

\subsection{Predictions at test time with multiple augmentations}
In \cref{sec:app:additional_results} we discussed that standard \gls{tta} ensembles the probabilities of the different test-time augmentations for the same input to make a prediction (\cref{eq:app:tta_avgprobs}). Because \losslog ensembles the logits, we also dissed ensembling the logits as test-time as an alternative (\cref{eq:app:tta_avglogits}). For completeness, we include these numbers in some figures in this appendix. While performance varies between datasets and models, we typically find that performance is comparable, though standard \gls{tta} that ensembles the probabilities typically has a slight edge.

\subsection{Extended results on CIFAR-100 and ImageNet}

In the remainder of this appendix, we show additional results for the datasets and networks considered in the main paper.

\begin{table*}[ht]
	\centering
	\small
	\begin{tabular}{llc | CCCC|CCCC}
		\toprule 
		\multirow{2}{*}{\textbf{Dataset}} & \multirow{2}{*}{\textbf{Network}} & \multirow{2}{1.cm}{\textbf{regula-rised?}} & \multicolumn{4}{c|}{\textbf{evaluation: central crop ($\ktest=1)$}} & \multicolumn{4}{c}{\textbf{evaluation: TTA ($\ktest=K_\text{max}$)}} \\
		&&& \ktrain=1 &  \lossstd & \lossprob & \losslog & \ktrain=1 & \lossstd & \lossprob & \losslog \\\midrule
		CIFAR-100 &WRN 16-4 & \xmark & 0.741 & 0.750 & 0.725 & 0.643 & 0.759 & 0.768 & 0.748 & 0.715\\
		CIFAR-100 &WRN 16-4 & \cmark & 0.741 & 0.759 &  0.758 &  0.759 & 0.759 & 0.772 & 0.772 & 0.772 \\\midrule
		ImageNet &NFResNet 101 & \xmark & 0.782 & 0.793 & 0.766 & 0.696 & 0.801 & 0.808 & 0.786 & 0.757 \\
		ImageNet &NFResNet 101 & \cmark & 0.782 & 0.801 & 0.804 & 0.804 & 0.801 & 0.806 &0.811 & 0.810 \\ \midrule
		ImageNet & NFResNet 50 & \xmark & 0.773 & 0.781 & 0.760 & 0.695 & 0.793 & 0.799 & 0.782 & 0.753  \\
		ImageNet & NFResNet 50 & \cmark & 0.773 & 0.784 & 0.789 & 0.787 & 0.793 & 0.792 & 0.797 & 0.797\\
		\bottomrule
	\end{tabular}
	\vskip-0.2em
	\caption{Top-1 test accuracy generalisation performance %
	for all three losses with and without regulariser evaluated on central crops and with test-time augmentations. For the losses we use $\ktrain=16$ on CIFAR and $\ktrain=8$ on ImageNet.
	}
	\label{tab:app:results}
\end{table*}

\begin{figure*}[htb]
    \centering
    \includestandalone[mode=image]{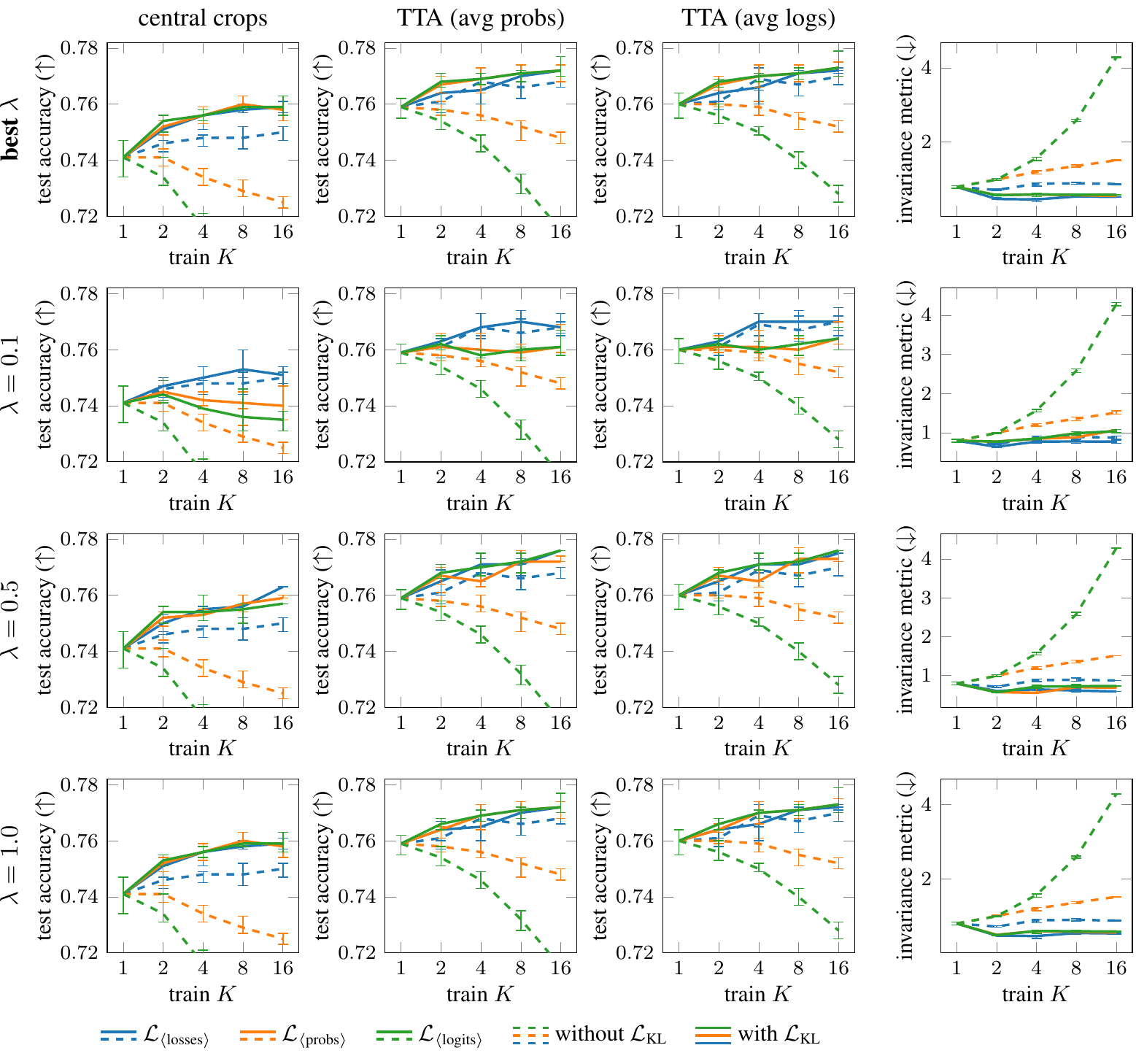}
    \caption{Extended results for CIFAR-100 with a WideResNet 16-4. We plot mean values over $5$ reruns and errorbars denote the minimum and maximum value to indicate spread between runs.}
    \label{fig:app:cifar_wrn}
\end{figure*}

\clearpage
\paragraph{CIFAR-100 with a WideResNet 16-4} 
\cref{fig:app:cifar_wrn}

\begin{figure*}[htb]
    \centering
    \includestandalone[mode=image]{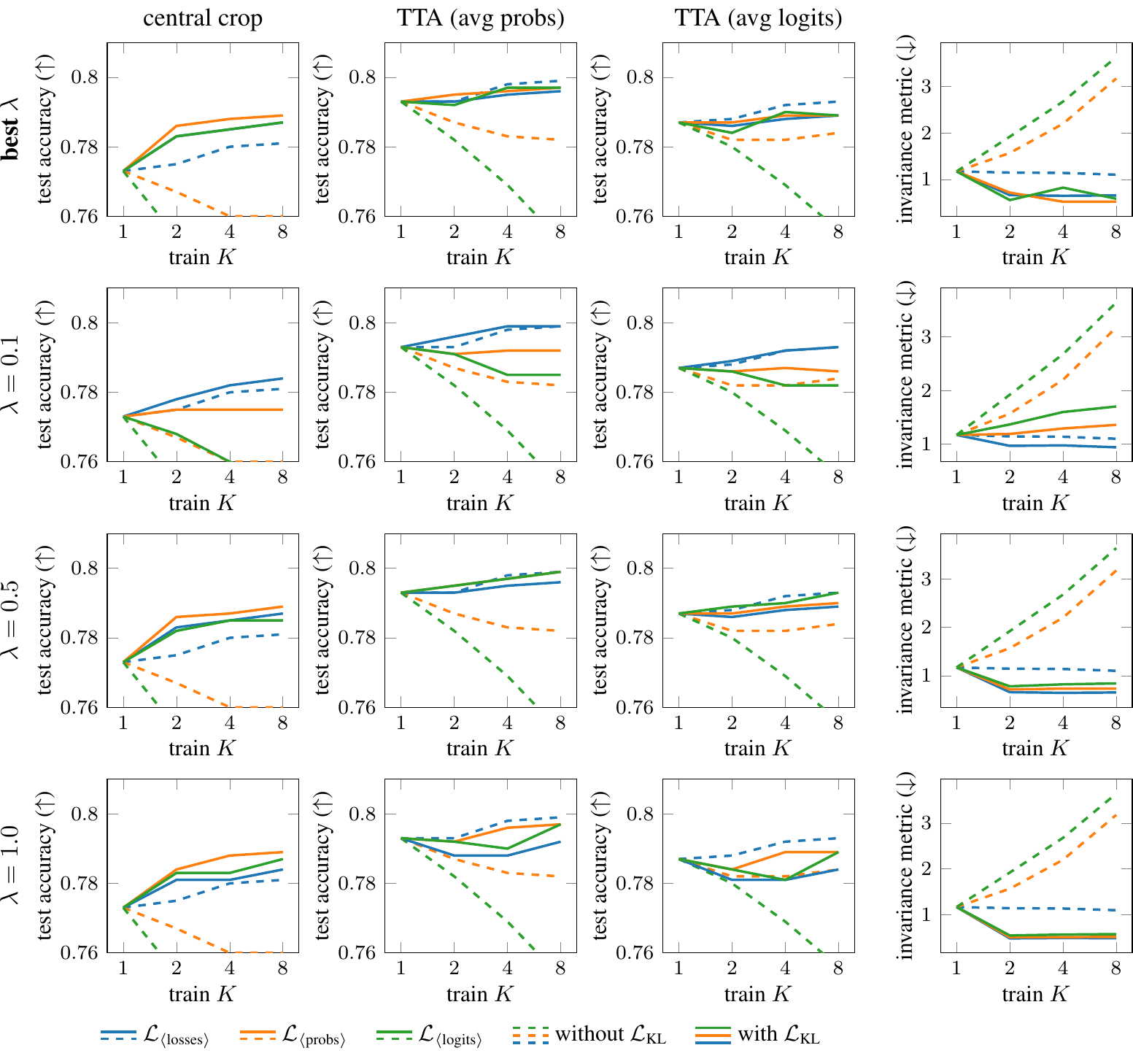}
    \caption{Extended results for ImageNet with an NF-ResNet-50}
    \label{fig:app:imagenet_resnet50}
\end{figure*}

\clearpage
\paragraph{Imagenet with an NF-ResNet-50}
\cref{fig:app:imagenet_resnet50}

\begin{figure*}[htb]
    \centering
    \includestandalone[mode=image]{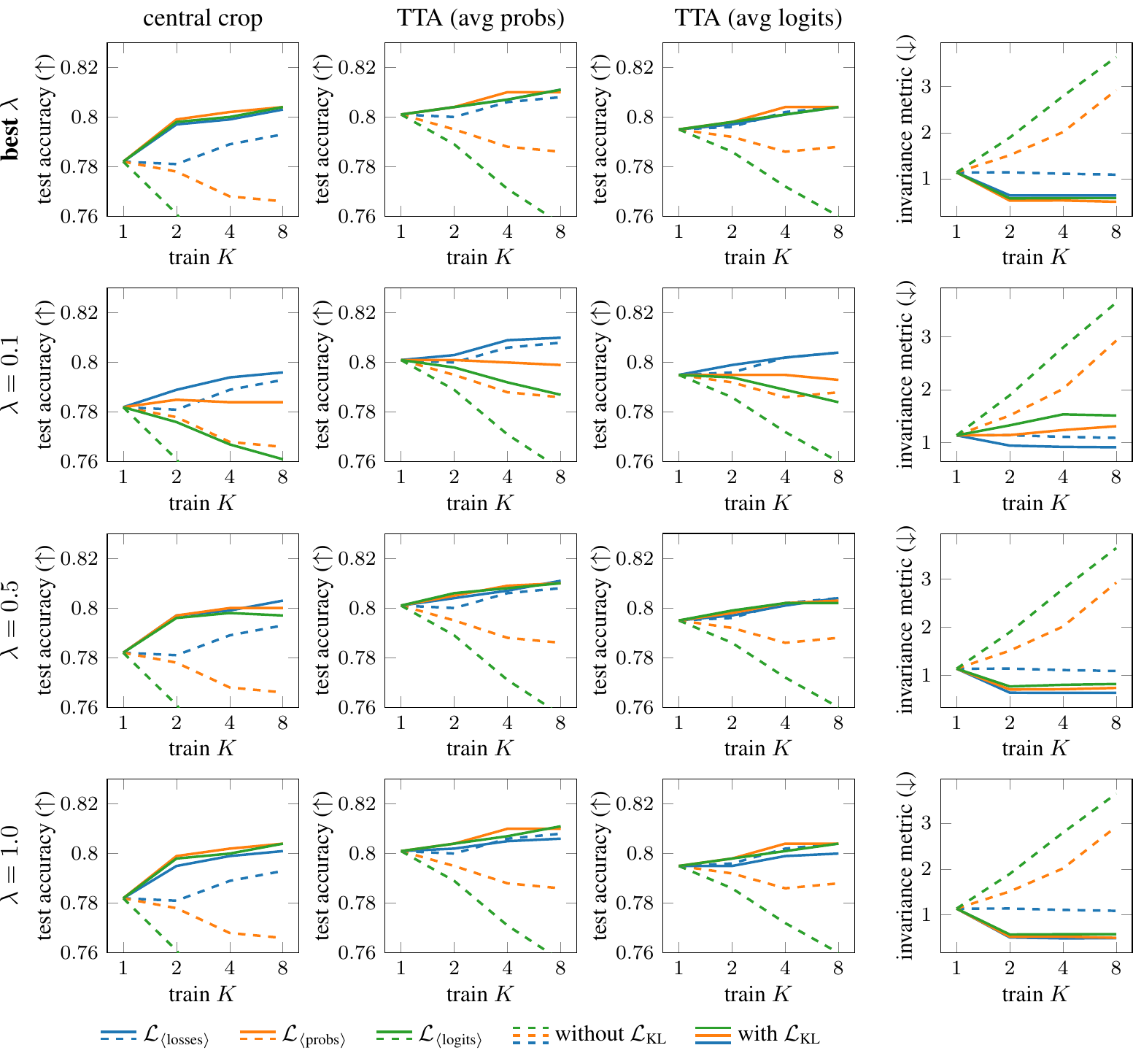}
    \caption{Extended results for ImageNet with an NF-ResNet-101}
    \label{fig:app:imagenet_resnet101}
\end{figure*}

\clearpage
\paragraph{Imagenet with an NF-ResNet-101}
\cref{fig:app:imagenet_resnet101}

\begin{figure*}[htb]
    \centering
    \includestandalone[mode=image]{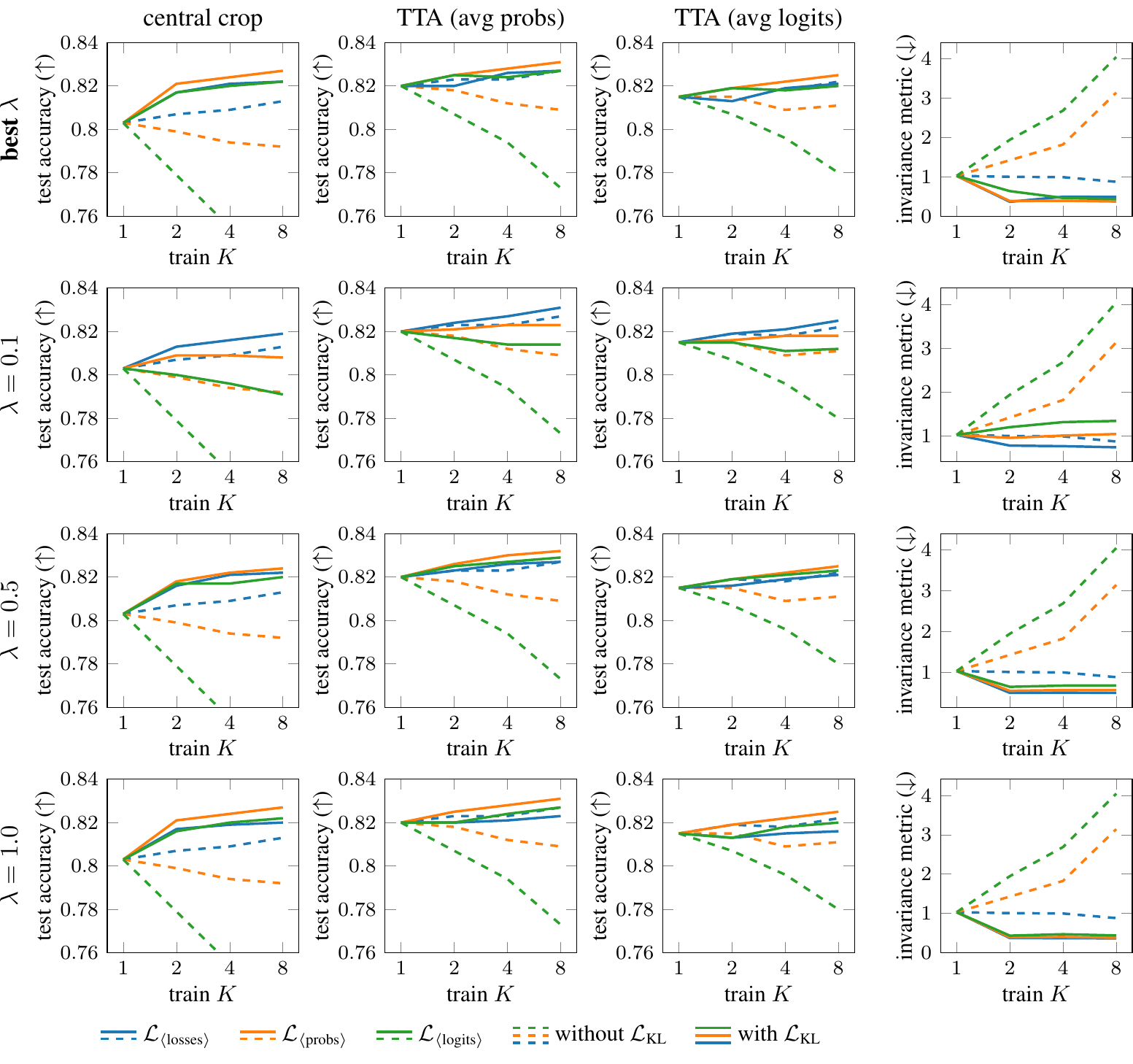}
    \caption{Extended results for ImageNet with an NFNet-F0}
    \label{fig:app:imagenet_nfnet}
\end{figure*}

\clearpage
\paragraph{Imagenet with an NFNet-F0}
(\cref{fig:app:imagenet_nfnet})
The NFNet-F0 uses stronger RandAugment \citep{randAugment} augmentations both during training and for TTA. As discussed previously in \cref{sec:app:experimental_details}, these augmentations can so strongly distort the images, that using it for TTA may not be most appropriate. For completeness, we show the results for TTA, though they should be viewed with a grain of salt.

\end{document}